\documentclass[10pt,conference]{ieeeconf} 

\IEEEoverridecommandlockouts                              
\usepackage[OT1]{fontenc} 
\usepackage{cite}
\usepackage{float}
\usepackage{amsmath}
\usepackage{dsfont}
\usepackage{amsfonts}
\usepackage{graphicx}

\usepackage{booktabs}
\usepackage{placeins}
\usepackage{xr}
\usepackage{bbding}
\usepackage{array}

\usepackage{relsize}
\usepackage{graphicx}
\usepackage{esint}
\usepackage[normalem]{ulem}
\usepackage{arydshln}
\usepackage{eurosym}
\usepackage{color}
\usepackage{amsmath,amssymb}
\usepackage{multirow}
\usepackage{textcomp}
\usepackage{tabularx}
\usepackage{placeins}
\usepackage{lscape}

\usepackage{comment}
\usepackage{mathrsfs}

\usepackage{algorithm}
\usepackage{algpseudocode}

\usepackage{hyperref}       
\usepackage[dvipsnames]{xcolor}
\usepackage{colortbl}
\usepackage{graphicx}

\usepackage{amsmath}

\usepackage{subfigure}
\usepackage{capt-of}
\usepackage{makecell}

\usepackage{mathtools}
\usepackage{pifont}

\title{\LARGE \bf
The Research of Group Re-identification from Multiple Cameras
}

\author{{Hao Xiao}
}

\begin{document}

\maketitle
\thispagestyle{empty}
\pagestyle{empty}

\begin{abstract}

Object re-identification is of increasing importance in visual surveillance. Most existing works focus on re-identify individual from multiple cameras while the application of group re-identification (Re-ID) is rarely discussed. We redefine Group Re-identification as a process which includes pedestrian detection, feature extraction, graph model construction, and graph matching. Group re-identification is very challenging since it is not only interfered by view-point and human pose variations in the traditional re-identification tasks, but also suffered from the challenges in group layout change and group member variation. To address the above challenges, this paper introduces a novel approach which leverages the multi-granularity information inside groups to facilitate group re-identification. We first introduce a multi-granularity Re-ID process, which derives features for multi-granularity objects (people/people-subgroups) in a group and iteratively evaluates their importances during group Re-ID, so as to handle group-wise misalignments due to viewpoint change and group dynamics. We further introduce a multi-order matching scheme. It adaptively selects representative people/people-subgroups in each group and integrates the multi-granularity information from these people/people-subgroups to obtain group-wise matching, hence achieving a more reliable matching score between groups. Experimental results on various datasets demonstrate the effectiveness of our approach.

\end{abstract}

\section{Introduction}
\label{sec:intro}

\begin{figure}[t!]
  \centering
  \includegraphics[width=0.4\textwidth]{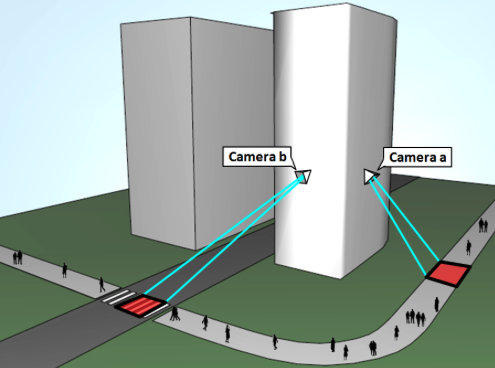}
  \caption{Vitual scenarios of multiple cameras surveillance}
    \label{fig:camera}
\end{figure}

Object re-identification (Re-ID), which aims at identifying a specific object indicated by a probe image from a set of gallery images captured from cross-view cameras, has been studied for years \cite{xiao2018group, lin2019group}. It has been applied in many industries to help with tracking and motion prediction \cite{tang2018single, gan2023mgtr, zhang2023sapi}. Fig.~\ref{fig:camera} shows a vitual scenarios of multiple cameras surveillance. A lot of researches have studied on object re-identification, where many of them focus on developing more reliable feature representation \cite{SalientColor, rankboost,3,4,OTF}, deriving more accurate feature-wise distance metric \cite{1,kernel-based,prid,mirror,zhang2016learning,chen2015relevance}, and handling local spatial misalignment between people \cite{2,tan2016dense,zhaorui1,zhaorui2,10,12,SLEPK,shenyang}. However, most existing works focus on the re-identification of individual object, while the re-identification of group of objects are seldom studied.

\begin{figure}[t!]
  \centering
  \subfigure[]{\includegraphics[width=2cm,height=4cm]{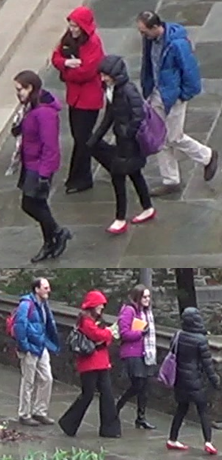}
    \label{fig1a}}
  \hspace{-1.6mm}
  \subfigure[]{\includegraphics[width=2cm,height=4cm]{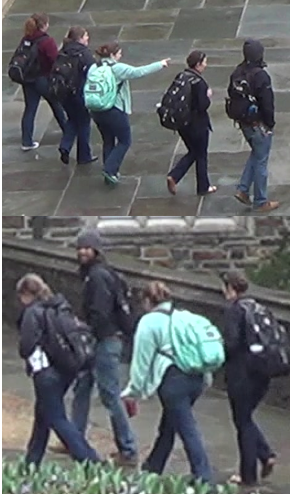}
    \label{fig1b}}
  \hspace{-1.6mm}
  \subfigure[]{\includegraphics[width=2cm,height=4cm]{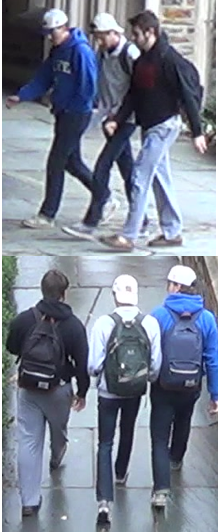}
    \label{fig1c}}
  \caption{Example of some challenging true group pair. (a): A true group pair with large layout variation. (b): A true group pair with different object number. (c): A true group pair  with large viewpoint change. (Best viewed)}
  \label{fig1}
  \vspace{-5mm}
\end{figure}

Basically, group re-identification has three challenging points:
\begin{enumerate}
\item[1)] The layout of individual object in group may include large change in different cameras (cf. Fig.~\ref{fig1a}).
\item[2)] People may often join or leave a group, which makes a group to include different objects in different views (cf. Fig.~\ref{fig1b}).
\item[3)] Group re-identification also share the same challenging points of individual person re-identification, such as viewpoint change and spatial misalignment (cf. Fig.~\ref{fig1c}).
\end{enumerate}

Although some methods \cite{zheng,cai} are developed to re-identify groups, they only simply utilize global feature of the entire group to perform group re-identification. Thus, They have limitations in creating reliable re-identification result in complex scenarios with large viewpoint change and group layout change across different cameras.

In this paper, we introduce the idea of \emph{group granularity}, where a coarser group granularity refer to  subgroups which including relatively large number of objects and a finer group granularity  contains fewer objects. we argue that multi-granularity information within a group is important in addressing the challenging point in group re-identification problem. For example, in Fig.~\ref{fig1a}, due to large camera viewpoint change and layout variation, the same group show large appearance difference in different cameras. If we simply extract global features for the entire group, the re-identification performance will be easily confused by visually similar but unrelated group. However, if we include information of finer group granularity(e.g. individual object), we are able to reduce the interference from global appearance and obtain more reliable re-identification result.

\begin{figure*}[t!]
  \centering
  \subfigure[]{\includegraphics[width=6cm]{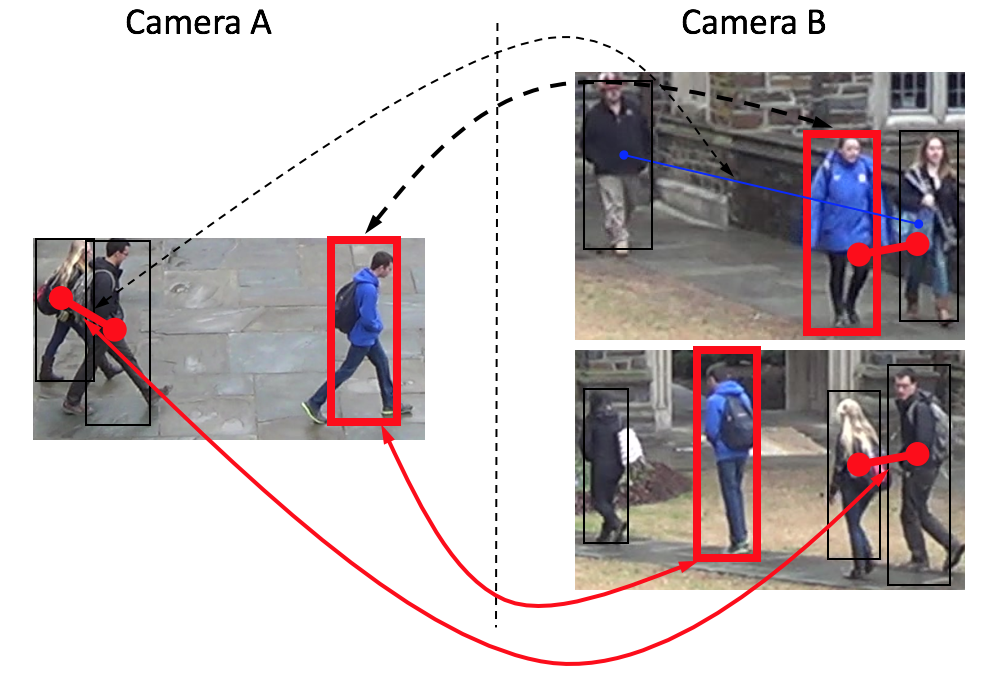}
    \label{fig2a}}
  \hspace{10mm}
  \subfigure[]{\includegraphics[width=6cm]{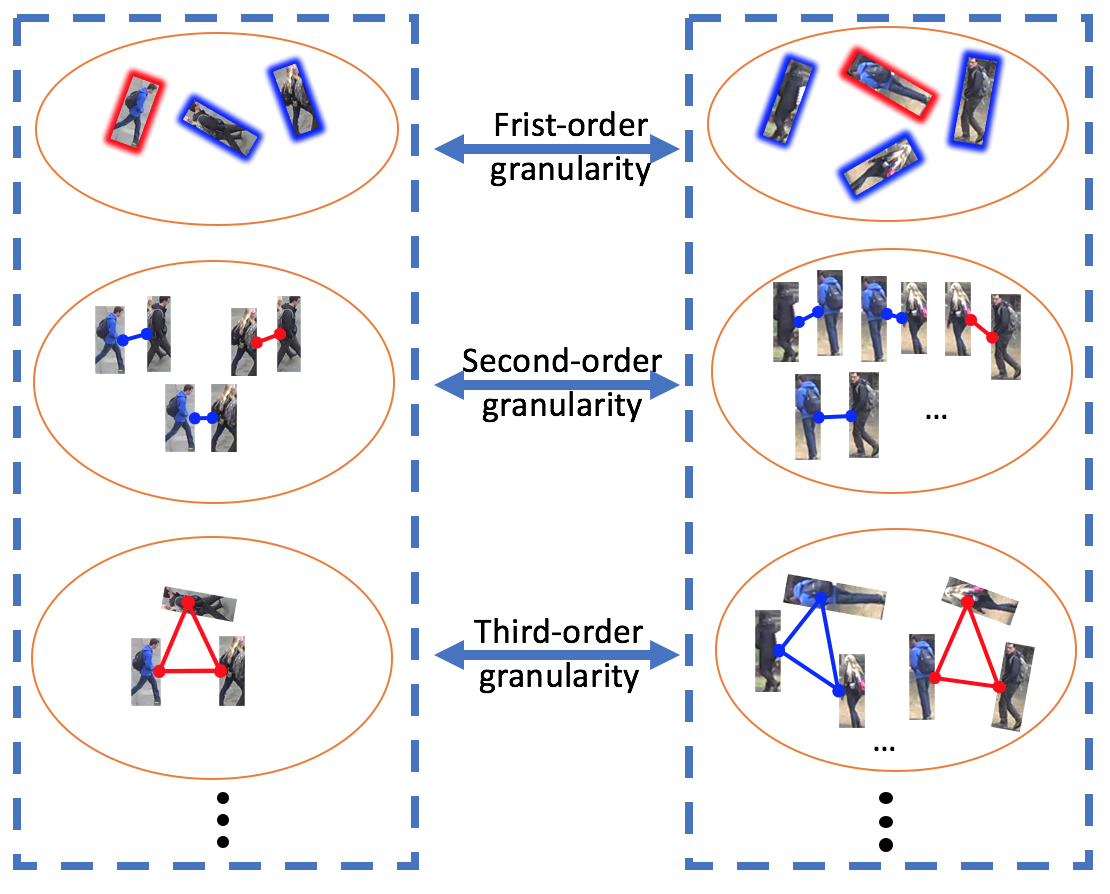}
    \label{fig2b}}
  \caption{(a) 2 visually similar gallery image in camera B to the same probe image in camera A. The true pair will be easily confused if we only consider individual or global object. If we consider middle-level group granularity of two-people subgroup and assign different importance of each people/people-subgroups, we are able to obtain a more accurate group re-identification result. (The red color denotes high value of importance and the blue color shows low value of importance) (b) Illustration of different importance of multi-granularity. (Best viewed in color)
  }
    \label{fig2}
    \vspace{-5mm}
\end{figure*}

Moreover, In Fig.~\ref{fig2a},  since 2 group images in camera B includes visually similar object to the probe image in camera A, the re-identification result will be easily confused if we only use individual or global object. Since the same group may maintain local layout structure in different camera views, the interference can be reduced by introducing middle-level group granularity (e.g. object subgroup of two people) or even coarser-level group granularity (e.g. object subgroup of three people) information to capture the subgroup layout similarity between groups.

Based on the above example, it is expected that the group re-identification performance will be obviously improved if we extract information from multi-granularity and integrate them (cf. Fig~\ref{fig2b}). At the same time, since groups in different cameras may include different objects (e.g. some objects leave/join group, or some objects are visually dissimilar, cf. Fig.~\ref{fig2a}) and different people/people-subgroups may carry different amount of information (e.g. people with blue jacket is more informative than people with black jacket). The group re-identification can be further facilitated if the importance of people/people-subgroups in each granularity can be properly evaluated and differentiated (cf. Fig.~\ref{fig2b}).

In summary, our contributions to re-identification are three folds.
\begin{enumerate}
\item[1)] We develop multi-granularity with different people/people-subgroups to obtain rich information to represent a group, and iteratively calculates and updates the importance of different people/people-subgroups in multiple granularity to handle the three challenging points of group re-identification.
\item[2)] Under this framework, we propose a multi-order matching process which adaptively picking up informative people/people-subgroups and integrating multi-granularity information from these people/people-subgroups to perform group-wise matching.
\item[3)] We construct a new and challenging dataset which includes large variation of camera view and group layout.
\end{enumerate}

\section{Related work}

The problem of person re-identification has attracted many researchers' attentions. Early works focus on developing proper handcraft features \cite{SalientColor, rankboost,3,4,OTF} or distance metrics  \cite{1,kernel-based,prid,mirror,zhang2016learning,chen2015relevance} to represent the similarity between cross-camera objects. Recently, due to the effectiveness of deep neural networks in learning discriminative features, many deep re-identification are proposed which utilized deep neural networks to enhance the reliability of feature representations or similarity metrics \cite{deep1,deep2}. Since most of these works do not effectively model the spatial misalignment among local patches inside images, their performances are still impacted by the interferences from viewpoint or human-pose changes. 

In order to address the spatial misalignment problem, some patch-based methods are proposed \cite{2,tan2016dense,10,12,SLEPK,shenyang} which decompose images into patches and perform an online patch-level matching to exclude patch-wise misalignments. In \cite{shenyang}, Shen et al. proposed a boosting-based approach to learn a correspondence structure which indicates the patch-wise matching probabilities between images from a target camera pair. The learned correspondence structure can capture the spatial correspondence pattern between cameras but also handle the viewpoint or human-pose variation in individual images. Then a global-based matching process was introduces which integrates a global matching constraint over the learned correspondence structure to exclude cross-view misalignments during the image patch matching process.

Besides, other researchers also further strengthen the capability of person re-identification by extending re-identification algorithm to other object types (e.g. cars \cite{car-reid}) or applying to more complex scenarios (e.g. multiple cameras \cite{multi-camera}, long object sequence \cite{zhengliang1,zhengliang2}, real-scene and untrimmed video \cite{end-to-end,zheng2016person,xiao2017ian}). In \cite{multi-camera}, Das et al. achieved a consistent and optimal re-identification by posing the problem of re-identification as an optimization problem that minimizes the global cost of associating pairs of targets on the entire camera network constrained by a set of consistency criteria. Zheng et al. \cite{zhengliang1} considers person re-identification in videos and introduced four re-identification strategies: image-to-image, image-to-video, video-to-image, and video-to-video, where the first mode is mostly studied in literature. The video-to-video pattern is where Zheng focused on since both probe and gallery units contain much richer visual information than single images. Xiao et al. \cite{end-to-end} considered re-identifying person in real-world scenarios where the annotations of pedestrian bounding boxes are unavailable and the target person needs to be searched from a gallery of whole scene images. Xiao proposed a new deep learning framework for person search which jointly handle both aspects in a single convolutional neural network Instead of breaking it down into two separate tasks: pedestrian detection and person re-identification.

Most of the existing works focus on the re-identification of individual object. They can not be easily extended to re-identify groups due to the layout change cross the camera. Only a few works are developed to address the problem of group re-identification. For example, Zheng et al. \cite{zheng} used a fixed template to segment group image into multiple ring regions and blocks, then proposed the Center Rectangular Ring Ratio-Occurrence Descriptor (CRRRO) and Block based Ratio-Occurrence Descriptor (BRO) to capture the the ratio information within and between regions. It is assumed that local variations of visual appearance in a ring region are stable from different cameras. Cai et al. \cite{cai} further proposed a discriminative covariance descriptor to capture both appearance and statistic properties of image region to maintain illumination invariant. However, both of these works focus on deriving global features to represent the overall characteristic of a group image/image region. They all have limitation in handling groups with large group layout variation in complex scenario (cf. Fig.~\ref{fig1}). Instead of design global feature, Zhu et al. \cite{zhu} formulate group re-identification as a patch matching based framework. Discriminative salience channels are learned to filter out highly unreliable and non-informative patch matches between two group images, while retain true matches undergoing appearance variations. The resulting candidate correspondences are further explored by a consistent matching process, which prefers coherent matches in true group image pairs. However, the problem of group layout change and group member variation are still not well-handled in their approach.

Although, some methods \cite{group-info1,group-info2} introduce local information of object pairs within the same camera to facilitate person re-identification process. They only use pair information to improving the matching accuracy of individual while the characteristic of groups are less considered and not fully modeled. For example, Assari et al. \cite{group-info1} considered a re-identification task in crowded scenes and modeled multiple personal, social and environmental (PSE) constraints on human motion across cameras. The personal constraints include appearance and preferred speed of each individual assumed to be similar across the non-overlapping cameras. The social constraints are quadratic in nature, i.e. occur between pairs of individuals, and modeled through grouping and collision avoidance. Finally, the environmental constraints capture the transition probabilities between gates (entrances / exits) in different cameras, defined as multi-modal distributions of transition time and destination between all pairs of gates. Assari incorporated these constraints into an energy minimization framework for solving individual person re-identification.

Besides, saliency based method \cite{zhaorui1,zhaorui2,bak1,bak2} estimate the saliency distribution relationship between images and utilize it to control the patch-wise matching process so as to obtain a best patch-wise matching between images to avoid spatial misalignment due to viewpoint change. Following the similar line, Bak et al. \cite{bak1} introduced a hand-crafted Epanechnikov kernel to determine the weight distribution of patches. Bak and Carr \cite{bak2} further introduced a deformable model to obtain a set of weights to guide the patch matching process. However, our approach also differ form these in :
(1) These methods simply model the feature and consider patch-wise matching in a single granularity. In contrast, our approach introduce multi-granularity to fully capture the characteristic of a group.
(2) These methods are only applied to individual object where patches in each image have relatively organized structure (e.g. patches for head always in the upper part of image, while patches for foot always in the bottom). Comparatively, out approach mainly focus on a more challenging task of group re-identification where the layout of object in a group has large variations and irregular.

Since our group re-identification approach includes finding one-to-one mapping between two groups, another thread of related works is graph matching which is an essential problem in computer vision. it is related to various research areas in computer vision, pattern recognition, and machine learning. The problem of graph matching is to determine a mapping between the nodes of the two graphs that preserves the relationships between the nodes as much as possible. In computer vision, it is widely known that the fundamental problem of establishing correspondences between two sets of visual features can be effectively solved by graph matching. Thus, graph matching is used in various tasks, such as feature tracking, image retrieval, object recognition, and shape matching. It is challenging to find perfect correspondences due to various reasons: imperfect feature descriptor, deformation in feature location due to viewpoint change or class variation, and outlier problem. The correspondence problem is well formulated as graph matching. Each graph is constructed with nodes representing features, and edges describing relations between two features. Correspondences are established by determining the mapping between two graphs, which preserves as much attributes as possible. Since the graph matching task is known to be NP-hard, approximate solutions are required. Cho et al. \cite{RRWM} introduced a random walk view on the problem and propose a robust graph matching algorithm against outliers and deformation. Matching between two graphs is formulated as node selection on an association graph whose nodes represent candidate correspondences between the two graphs. The solution is obtained by simulating random walks with reweighting jumps enforcing the matching constraints on the association graph. Some researchers believe that pairwise relations are not enough to incorporate the information about the entire geometrical structure of features. So several researchers have tried to embed higher-order information into their problem formulations to overcome the limitation of pair-wise similarity. For example, Lee et al. \cite{RRWHM} generalized the previous hyper-graph matching formulations to cover relations of features in arbitrary orders, and propose a novel algorithm by reinterpreting the random walk concept on the hyper-graph in a probabilistic manner. In Lee's formulation, relations in different orders are embedded altogether in a recursive manner, yielding a single higher-order similarity tensor. comparatively, our approach separate the information of different orders into different layer and consider two types of affinity which are inter-order affinity and intra-order affinity. Thus, information of multi-granularity can be fully modeled.

\section{Method}

This section first introduce the details of our multi-granularity group re-identification framework. Second, pedestrian detection algorithm is illustrated. Third, we introduce the concept of multi-granularity and show the scheme of computing the multi-granularity features. Then, in order to learn an importance value in an unsupervised manner, an importance evaluation module is addressed which integrate three factors: Saliency, Purity, Stability. Finally, an iterative process is introduced, which adaptively calculates and updates the importance of different people/people-subgroups in multiple granularity and improve the performance iteratively.

\begin{figure}[t!]
  \centering
  \includegraphics[width=0.5\textwidth]{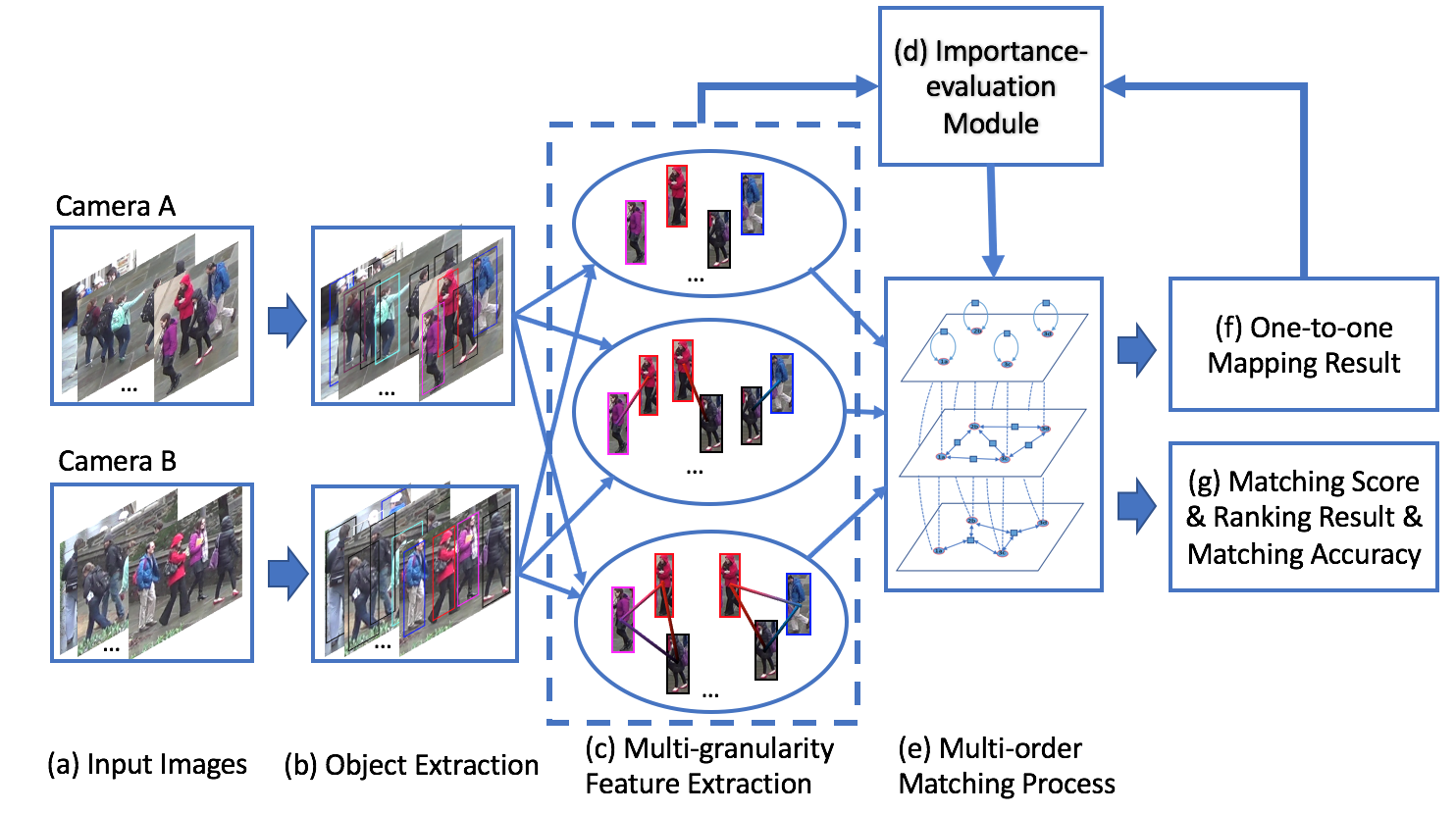}
  \caption{Framework of the proposed approach.}
    \label{fig:framework}
    \vspace{-5mm}
\end{figure}

\subsection{Overview}\label{section:overview}

The framework of our approach is shown in Fig.~\ref{fig:framework}.
Firstly, given two set of group images, we identify the objects in each group and then extract the feature of every objects. Here we combine RGB, HSV, YCbCr, Lab, YIQ and 16 Gabor texture features to get color and texture information of each individual. The combination of both kinds of feature effectively extract both the texture information and color information for every individual, which is very essential for the similarity measurement. Then, middle-granularity and coarse-granularity object attributes are defined by concatenating individual descriptor and relative distance/angle histogram \cite{harg} which are used to reflect the spatial relationship among multiple pedestrians.

Secondly, an importance evaluation module is applied, which adaptively analyzes statistics and correlation of multi-granularity features in probe and gallery images and evaluates the importance. This is based on the assumption that different multi-granularity objects may have different contribution to this group. Thus, it is necessary to assign different value of importance to different people/people-subgroup in group re-identification. we consider three factors to evaluate the importance of an individual which are saliency, purity and stability. And then, based on the importance of individual, the importance of people-subgroup is also introduced.

Finally, the multi-granularity features of probe and gallery images will be input into multi-order matching process under the guidance of importances of people/people-subgroups. The one-to-one matching result will be further used to update the importance of people/people-subgroups. Thus, matching result will also change accordingly. This process is repeated for several iterations until the one-to-one matching result become convergence or reach the maximum iteration number. Finally, the group re-identification result is achieved by ranking gallery images to every probe image according to their matching scores.

\subsection{Pedestrian Detection\label{section:detection}}

\begin{figure}[t!]
  \centering
  \includegraphics[width=0.5\textwidth]{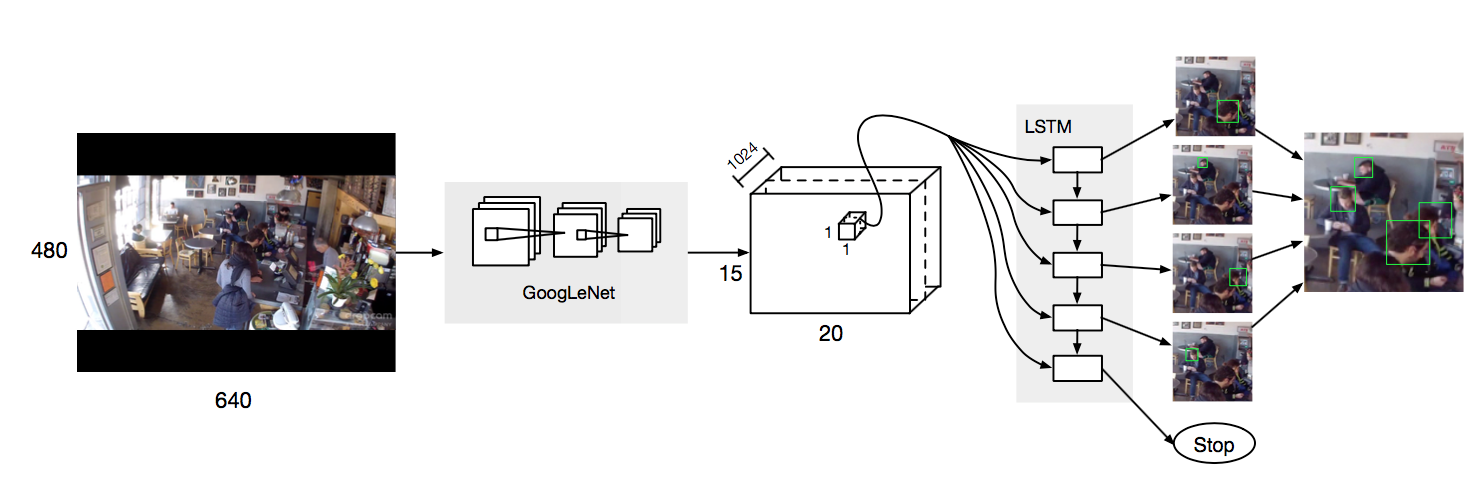}
  \caption{Framework of pedestrian detection \cite{reinspect}}
    \label{fig:detection}
    \vspace{-5mm}
\end{figure}

Detection and segmentation via images undergoes active research \cite{huang2024duospacenet, li2022bevformer, reinspect, googlenet}. Stewart et al. \cite{reinspect} proposed a new architecture for detecting objects in images which is an end-to-end method that accepts images as input and generates a set of object bounding boxes as output directly. It demands both distinguishing objects from the background and correctly estimating the number of distinct objects and their locations. Such an end-to-end approach capable of directly outputting predictions would be advantageous over methods that first generate a set of bounding boxes, evaluate them with a classifier, and then perform some form of merging or non-maximum suppression on an over complete set of detections. 

The detection process is show in Fig.~\ref{fig:detection} which first encodes an image into high level descriptors via a convolutional network (GoogLeNet \cite{googlenet}), and then decodes that representation into a set of bounding boxes. Follow with a recurring network of LSTM units which is the main part for predicting variable length output. It transform each image into a grid of 1024 dimensional feature descriptors at striped regions throughout the image. The 1024 dimensional vector summarizes the contents of the region and carries rich information regarding the positions of objects. The LSTM draws from this information source and acts as a controller in the decoding of a region. At each step, the LSTM outputs a new bounding box $b_{pos}$ and a corresponding confidence $b_c\in[0,1]$ that a previously undetected person will be found at that location. Boxes are encouraged to be produced in order of descending confidence. When the LSTM is unable to find another box in the region with a confidence above a pre-specified threshold (e.g. 0.5), a stop symbol is produced. The sequence of outputs is collected and presented as a final description of all object instances in the region.

Denote the corresponding set of ground truth bounding boxes as $G=\{b_{pos}^i|i=1,\dots,M\}$, and the set of candidate bounding boxes generated by the model as $C=\{ \tilde{b}_{pos}^j | j=1,\dots,N\}$. Hungarian algorithm is used to return an injective function $g:G\rightarrow C$, i.e. $f(i)$ is the index of candidate hypothesis assigned to ground-truth hypothesis. Given f, a loss function is defined as follows:

\begin{align}
  L(G,C,f)=\alpha \sum_{i=1}^{|G|}||(b_{pos}^i,\tilde{b}_{pos}^{f(i)}||_1+ \nonumber \\
  \sum_{j=1}^{|C|}l_c(\tilde{b}_c^j,\mathds{1}\{f^{-1}(j)\neq \empty \varnothing \})
\end{align}

where $l_c$ is a cross-entropy loss on a candidate's confidence that it would be matched to a ground-truth, $\tilde{b}_c^j$ is the confidence corresponding to  $\tilde{b}_{pos}^j$, and $\alpha$ is a term trading off between confidence errors and localization errors. In the experiment, we empirically set $\alpha=0.03$ for the best performance.

\begin{figure}[t!]
  \centering
  \includegraphics[width=0.5\textwidth]{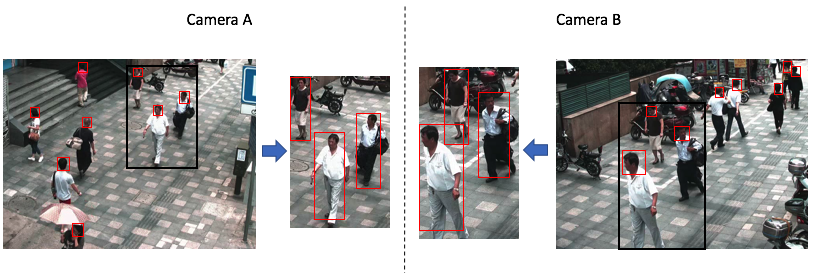}
  \caption{Example of pedestrian detection}
    \label{fig:example}
    \vspace{-5mm}
\end{figure}

We use a pre-trained model of \cite{reinspect} to identify heads in the origin image and crop individual people based on the size of head. The motivation for detecting only pedestrian heads comes from detection of indoor pedestrian or in crowded scenes, where the body of a person may be invisible. It will be more reliable to detect head instead of whole body when we perform group re-identification. Fig.~\ref{fig:example} shows a example of identifying people within a group according to the known group bounding box.

\subsection{Multi-granularity Feature Extraction
\label{section:feature_extract}}
In order to make fully use of the characteristic of a group, we need to consider multi-granularity structure.
After we apply pedestrian detection algorithm \cite{reinspect} to identify people in each group image, we derive multi-granularity feature accordingly. In the experiment, we construct three granularity structure which include finer-granularity (i.e. individual), middle granularity (i.e. two people subgroup), coarse granularity (i.e. three people subgroup).

For finer granularity, we adopt the feature setting in \cite{mirror} to get the feature of an individual person. We equally partitioned each image into 18 horizontal stripes, and RGB, HSV, YCbCr, Lab, YIQ and 16 Gabor texture features were extracted for each stripe. For each feature channel, a 16D histogram was extracted and then normalized by L1-norm. All histograms were concatenated together to form a single vector. The combination of both kinds of feature effectively extract both the color information and texture information for every image, which is very essential for the similarity measurement.

\begin{figure}[t!]
  \centering
  \includegraphics[width=0.5\textwidth]{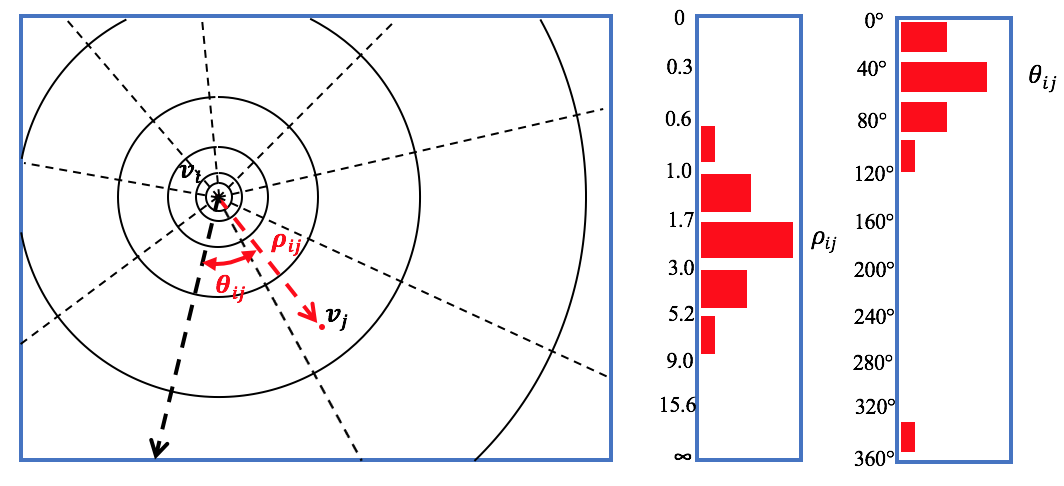}
  \caption{Histogram of log-polar bins for edge attributes. This attribute is a concatenation of log-distance and polar-angle histograms. Each histogram is represented by a discrete Gaussian window centered at a bin. The log-distance $\rho_{ij}$ of edge $e_{ij}$ is measured relative to the scale of group image. The polar-angle $\theta_{ij}$ of edge $e_{ij}$ is measured from the from the reference direction through vi (shown as a black thick arrow), which is the fitting line direction of this group by applying RANSAC algorithm to center points of all individual.}
  \label{fig:edge}
  \vspace{-5mm}
\end{figure}

For middle granularity, we need to consider the representation of length and angle of an edge. We use histogram-attributed relational graph (HARG, \cite{harg}), wherein all edge and angle attributes are represented by histogram distributions. As is illustrated in Fig.~\ref{fig:edge}, consider an edge $e_{ij}$ from node $v_i$ to node $v_j$. The vector between these two nodes can be expressed in polar coordinates as $(\rho_{ij},\theta_{ij})$. We transform this into a histogram-based attribute, which is invariant to the characteristic scale and orientation. For length, we use uniform bins of size $n_L$ in the log space with respect to the position and scale, making the histogram more sensitive to the position of nearby points. The log-distance histogram $L_{ij}$ is constructed on the bins by a discrete Gaussian histogram centered on the bin for $\rho_{ij}$:
\begin{align}
L_{ij}(k)=f_L(k-m),
\end{align}
$$
\text{s.t.} ~~ f_L(x)=\mathcal{N}(0,\sigma_L), ~~ \rho_{ij} \in \text{bin}_{\rho}(m)
$$

where $\mathcal{N}(\mu,\sigma)$ represent a discrete Gaussian window of size $\sigma$ centered on $\mu$, and $\text{bin}_{\rho}(k)$ denotes the kth log-distance bin from the center of $v_i$. For angle, we use uniform bins of size $2\pi / n_P$. The polar-angle histogram $P_{ij}$ is constructed on it in a similar way, except that a circular Gaussian histogram centered on the bin for $\theta{ij}$ with respect to the characteristic orientation of $v_i$, is used:
\begin{align}
P_{ij}(k)=f_P(k-m),
\end{align}
$$
\text{s.t.} f_P(x)=\mathcal{N}(0,\sigma_P)+\mathcal{N}(\pm n_P,\sigma_p), ~~ \theta_{ij} \in \text{bin}_{\rho}(m)
$$

where additional Gaussian terms in $f_P(x)$ induce the circular bins for angle. Thus, the final descriptor (or histogram) of middle granularity is composed by concatenating the descriptor of two related nodes $A_i,A_j$, the log-distance $L_{ij}$, and the polor-angle $P_{ij}$, which is $A_{i,j}=[A_i;A_j;L_{ij};P_{ij}]$.

For coarse-granularity, we consider the three internal angle $(\theta_1,\theta_2,\theta_3)$ of triangle $(v_i,v_j,v_k)$. Define the internal angle attribute as $I_{ijk}=[\sin(\theta_1);\sin(\theta_3);\sin(\theta_3)]$. Thus, the final descriptor (or histogram) of coarse granularity can defined by concatenating descriptors of three vertices, descriptors of three edges, and internal angle discriptor of three edge, which is $A_{i,j,k}=[A_i;A_j;A_k;L_{ij};P_{ij};L_{jk};P_{jk};L_{ki};P_{ki};I_{ijk}]$. In our experiment, we set $\sigma_{L}=\sigma_{P}=5,n_L=n_P=9$. 

Note that the approach to extract feature in different granularity is not limited. In practice, many off-the-shelf methods can be used.

\subsection{Importance Evaluation
\label{section:weight}}
Since each people/people-subgroups in a group may have different contribution to this group, it is necessary to assign different value of importance to different people/people-subgroups in group re-identification. In our framework, we consider three factors to evaluate the importance of an object which are saliency, purity and stability. In the following, we will explain these three factor and why they affect the performance of group re-identification. Then, based on the importance of individual person, the importance of people-subgroup is introduced.

\begin{figure}[t!]
  \centering
  \includegraphics[width=0.45\textwidth]{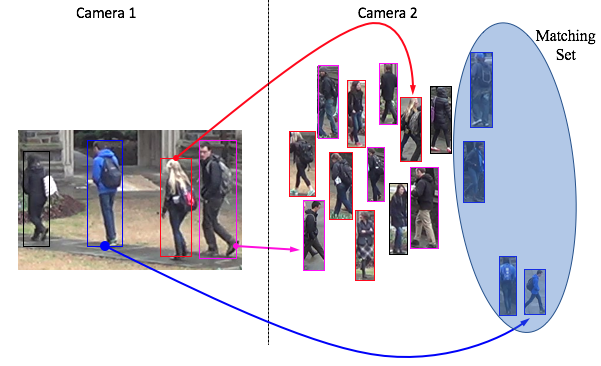}
  \caption{Illustration of saliency, purity and stability. Person with blue box has a relatively high value of importance since the matching set has only limited number of visually similar objects and the distribution of matching set is not mix up with other matching sets. People with red and purple box have a relatively low value of importance since the matching sets have many visually similar objects and these two matching sets is mix up with each other. People with black box has the minimum value of importance because the independent position in this group image. (Best viewed in color)}
    \label{fig:sal}
  \vspace{-5mm}
\end{figure}

{\bf Saliency.}
In \cite{zhaorui1}, salient patches are those possess uniqueness property among a specific set. In our paper, we borrow the idea of salience to describe how unique a object is in a specific group set. This is based on an intuition that salient objects in group re-identification provide valuable information since it is discriminative in finding the same group across different view. For example, in Fig.~\ref{fig:sal}, salient object (i.e.  the object with blue boxes) is distributed far way from other objects. The salient object will be informative in group re-identification and should have a large value of importance.

In order to quantify saliency, we made a slight change of K-nearest neighbor salience in \cite{zhaorui1}. The feature vector of an object is represented as $x_{n}^{\Theta,m}$, where $(m,n,\Theta)$ denotes the $n$-th object in $m$-th group in camera $\Theta$, $\Theta \in \{A,B\}$. Then, a group in camera A can be represented as $\mathcal{T}^{A}(m)=\{ x_{n}^{A,m} | n=1, \cdots, N_{m}^{A} \}$, where $N_{m}^{A}$ is the number of objects in $m$-th image in camera A.
Denote the number of group images in the reference set by $N_{r}$. We define a matching set of object $x_{n}^{A,m}$ as 
\begin{align}
\mathcal{M}(x_{n}^{A,m})=\{ x| \mathop{matching}_{\hat{x}\in \mathcal{T}^{B}(p)}  (x_{n}^{A,m},\hat{x})=1, p=1,\cdots N_{r} \}
\end{align}

where $matching(x_{n}^{A,m},x_{q}^{B,p})=1$ indicates that $x_{n}^{A,m}$ matched to $x_{q}^{B,p}$ according to our multi-order matching process which will be detailed in \ref{chapter:matching}. Then, we define salience score as:
\begin{align}
\label{equation:sal}
score_{sal}(x_{n}^{A,m}) = \frac{D_{k}(\mathcal{M}(x_{n}^{A,m}))}{|\mathcal{M}(x_{n}^{A,m})|}
\end{align}

where $D_{k}$ denotes the distance of the $k$-th nearest neighbor. If the distribution of the matching set well reflects the test scenario, the salient object (e.g. the object with blue box) can only find limited number of visually similar objects, while the distribution of the matching set is not concentrated as shown in Fig.~\ref{fig:sal}. Thus, $score_{sal}(x_{n}^{A,m})$ is expected to be large. Comparatively, the common object (e.g. the object with red and purple box) will find large number of visually similar object while the distribution of matching set will be concentrated. These kind of object should have a relatively low importance since it will introduce confusion in group re-identification, i.e. $score_{sal}(x_{n}^{A,m})$ should be low. The goal of salience detection for group re-identificatioin is to identify persons who have unique appearance. We assume that if a person has such unique appearance, more than half of the people in the matching set are dissimilar with him/her. With this assumption, $k=round(|\mathcal{M}(x_{n}^{A,m})|/2)$ is used in our experiment, where $round(a)$ denotes the nearest integer to number $a$.

{\bf Purity.} Purity indicates how chaotic between two matching sets. For example, in Fig.~\ref{fig:sal}, the red matching set and purple matching set have a lot of overlapping which indicates that these two person are hard to differentiated. This will lead to the misalignment of these two person. Thus, we need to assign a low value of importance to these two objects to reduce the interference of chaos. 

In order to quantify purity, we first calculate the Wasserstein-1 distance, or the earth-mover distance \cite{2000emd} of two matching set:
\begin{align}
EMD(\mathcal{M}(x_{n_{1}}^{A,m}),\mathcal{M}(x_{n_{2}}^{A,m}))= \mathop{\min}_{f_{ij}} \frac{\sum_{i,j}f_{ij}d_{ij}}{\sum_{i,j}f_{ij}}
\end{align}
$$
\text{s.t.}~~~i=1,\cdots, |\mathcal{M}(x_{n_{1}}^{A,m})|, j=1,\cdots, |\mathcal{M}(x_{n_{2}}^{A,m})|,
$$
$$
f_{ij} \geq 0, \mathop{\sum}_{j}f_{ij} \leq 1, \mathop{\sum}_{i}f_{ij} \leq 1,
$$
$$
\mathop{\sum}_{i,j}f_{ij}=min(|\mathcal{M}(x_{n_{1}}^{A,m})|,  |\mathcal{M}(x_{n_{2}}^{A,m})|)
$$
where $f_{ij}$ denotes the flows and $d_{ij}$ indicates the distance between $i$-th element in  $\mathcal{M}(x_{n_{1}}^{A,m})$ and $j$-th element in $\mathcal{M}(x_{n_{2}}^{A,m})$.

Thus, we define purity score as:
\begin{align}
\label{equation:pur}
score_{pur}(x_{n_{i}}^{A,m})=\mathop{\sum}_{j\neq i}EMD(\mathcal{M}(x_{n_{i}}^{A,m}),\mathcal{M}(x_{n_{j}}^{A,m}))
\end{align}

If a matching set is pure enough, the corresponding object will be more informative (e.g. the object with blue box in Fig.~\ref{fig:sal}). Thus, $score_{EMD}(x_{n}^{A,m})$ is expected to be large. In contrary, a small value of $score_{EMD}(x_{n}^{A,m})$ indicates that the matching set of corresponding object mix up with other matching set. If we consider these two people equally (i.e. both of these two people have importance of 1), they will easily find a wrong match through graph matching algorithm. Ir is necessary to assign a low value of importance to these two people to avoid misalignment due to camera view change and pose variation.

{\bf Stability.} 
In a real-scene group of people, some people will leave/join the group due to their own walking behavior. Thus, we should take this fact into consideration in group re-identification. Intuitively, a person at a relatively independent position is more likely to disengage from this group, while a person has a strong correlation with other people (i.e. short distance with other people) in the group is more likely to maintain the waking manner. We find local outlier factor \cite{lof} is appropriate for describe this characteristic of group.

Local outlier factor has been widely used in anomaly detection which is based on a concept of a local density. Define reachability distance as:
\begin{align}
reachability\text{-}distance_k(A,B)=max\{ k \text{-} dist(b), dist(a,b)\}
\end{align}
where $k \text{-} dist(b)$ is the distance of object b to the $k$-th nearest neighbor. In words, the reachability distance of an object A from B is the true distance of the two objects, but at least the k-distance of object B. Objects that belong to the k nearest neighbors of B are considered to be equally distant. The reason for this distance is to get more stable results. Note that this is not a distance in the mathematical definition, since it is not symmetric.

Then, the local reachability density of object $a$ is defined by
\begin{align}
lrd(a)=1/(\frac{\sum_{b \in N_{k}(a)} reachability\text{-}distance_k(A,B)}{|N_{k}(a)|})
\end{align}
where $N_{k}(a)$ is the set of $k$ nearest neighbors of object a in the same group image. Then, the local outlier factor can be calculate as:
\begin{align}
LOF(a)=\frac{\sum_{b \in N_{k}(a)} \frac{lrd(b)}{lrd(a)}}{|N_{k}(a)|}
\end{align}

which is the average local reachability density of the neighbors divided by the object's own local reachability density. A value of approximately 1 indicates that the object is comparable to its neighbors (and thus not an outlier). A value below 1 indicates a denser region (i.e. inlier), while values significantly larger than 1 indicate a sparse region (i.e. outlier).

In this paper, we borrow this idea to model the stability of a person in a group. We applied the center coordinate of each person in a group to the calculation of LOF, and $k=round(N_{m}^{A}/2)$ is used in our experiment. Then, we define the stability score as:
\begin{align}
\label{equation:stb}
score_{stb}(x_{n}^{{A,m}})=\frac{1}{LOF(v_{n}^{A,m})}
\end{align}
where $v_{n}^{A,m}$ denotes the center coordinate in the $m$-th group image of $n$-th object in camera $A$.
Then, a person with stability score lager than 1 indicates this object is more capable to maintain its own walking behavior and show up in the same group in another camera (e.g. the object with blue box in Fig.~\ref{fig:sal}).  Comparatively, an object with stability score below 1 means this object is much likely to disengage from this group (e.g. the object with black box in Fig.~\ref{fig:sal}).

{\bf people/people-subgroup Importance.} Based on above three factors, we define the importance of $n$-th object in $m$-th group of camera $A$ as:
\begin{align}
\label{equation:importance1}
Impo&rtance(x_{n}^{A,m})=\widehat{score}_{sal}(x_{n}^{A,m})\nonumber \\
& +\alpha_{pur} \widehat{score}_{pur}(x_{n}^{A,m}) + \alpha_{stb} \widehat{score}_{stb}(x_{n}^{A,m})
\end{align}
where $\widehat{score}(\cdot)$ is a normalize function since all scores have their respective ranges. $\alpha_{pur}=\alpha_{stb}=1$ is used in our experiment.

Based on the importance of individual people, we are able to calculate the importance of people-subgroup. We define the importance of k-people subgroup (i.e. $k$-th order granularity importance) in a recursive manner as follows:
\begin{align}
\label{equation:importance2}
Impo&rtance(x_{n_{1},\cdots,n_{k}}^{A,m})=\Omega_{k}(x_{n_{1},\cdots,n_{k}}^{A,m})\nonumber \\
&+\lambda^{(k-1)}\sum_{l=1}^{k}Importance(x_{\{n_{1},\cdots,n_{k}\}\backslash n_{l}}^{A,m})
\end{align}

where $\lambda^{(k)}$ represents the weighting factor and $\Omega_{k}(x_{n_{1},\cdots,n_{k}}^{A,m})$ denotes the stability of $k$-th level granularity of $k$ objects $(n_{1},\cdots,n_{k})$ in $m$-th group from camera $A$. Note the subscript are used to represent object indices.

Specifically, we define edge stability (i.e. middle level granularity) and hyper-edge stability (coarse level granularity).
Intuitively, the shorter a edge is, the more stable and informative the edge is. On contrary, edges with longer length will vary greatly in different cameras. So we define the stability of middle level granularity as:
\begin{align}
\Omega_{2}(x_{n_{1},n_{2}}^{A,m})=exp[\sigma_{r}\sqrt{w^{2}+h^{2}}/dist(v_{n_{1}}^{A,m},v_{n_{2}}^{A,m})]
\end{align}
where $w$ and $h$ are the width and height of the origin group image.
For hyper-edge stability, we believe the equilateral triangle structure is most stable. So we define the stability of coarse level granularity as:
\begin{align}
\Omega_{3}(x_{n_{1},n_{2},n_{3}}^{A,m})=exp[-\frac{1}{\sigma_{s}}\sum_{i=1}^{3}|sin(\theta_{i})-sin(\frac{\pi}{3})|]
\end{align}
where $\theta_{i}, i=1,2,3$ is three internal angle of triangle ($v_{n_{1}}^{A,m},v_{n_{2}}^{A,m},v_{n_{3}}^{A,m}$) and we empirically set $\sigma_{r}=\sigma_{s}=0.5$ for the best performance.

\subsection{Iterative Process\label{section:iter}}

\begin{algorithm}
   \caption{Multi-granularity Group re-identification Process}
   \label{algorithm:framework}
   {\bf Input}: A set of training probe images from camera A and their corresponding cross-view images from camera B. \\
   {\bf Output}: Importance of multi-granularity $\mathcal{W}$, matching score and matching result $x$.
   \begin{algorithmic}[1]
      \State Apply object detection method \cite{reinspect} to identify the object in each group.
      \State Get multi-granularity features $x_{n}^{\Theta,m},x_{n_{1},n_{2}}^{\Theta,m},x_{n_{1},n_{2},n_{3}}^{\Theta,m}$, where $\Theta \in \{A,B \}$.
      \State Assign the importance of all people/people-subgroups in different granularity to be 1.
      \Repeat
        \State iter $\gets$ iter+1.
        \State Utilize the multi-order matching process to obtain the one-to-one mapping result $\hat{x}$ which will be detailed in ~\ref{chapter:matching}.
        \State For object $x_{n}^{\Theta,m}$, get its matching set $\mathcal{M}(x_{n}^{\Theta,m})$ according to one-to-one mapping result.
        \State Calculate the score of saliency, purity and stability by Eq.~\ref{equation:sal}, Eq.~\ref{equation:pur}, Eq.~\ref{equation:stb}, respectively.
        \State Assign new importance of people/people-subgroups $\mathcal{W}$ in different granularity according to Eq.~\ref{equation:importance1} and Eq.~\ref{equation:importance2}.
      \Until {$\mathcal{W}$ converges or reach maximum iteration number.}
      \State Obtain matching score of every group-wise match candidate and evaluate the performance of group re-identification.
     \end{algorithmic}
\end{algorithm}

As mentioned in previous section, the importance of people/people-subgroups in different granularity are depend on the one-to-one matching result. However, the matching result can not be obtained before we actually perform matching. Therefore, In this paper, we utilize an iterative process which updates the importance and matching result iteratively. Specifically, we first assume all importance to be 1 and apply multi-order matching process as initial match result. Then, we can use the initial matching result to update the importance of each people/people-subgroups in a group and apply another time of multi-order matching process. This procedure is repeat for several iterations until the one-to-one matching result become convergence or reach the maximum iteration number. The entire multi-granularity group re-identification process can be describe in Algorithm~\ref{algorithm:framework}.

Besides, although the exact convergence of our learning process is difficult to analyze due to the matching result and importance of each people/people-subgroups is interactive, our experiments show that most the one-to-one matching result become stable within 5 iterations, which implies the reliability of our approach.

\section{Multi-order Matching Module}
\label{chapter:matching}
After obtain Multi-granularity features, it is important to perform matching to obtain 1) one-to-one matching result and 2) matching score. In this section, we first develop a multi-order matching process based on random walking \cite{RRWM,RRWHM} and obtain one-to-one mapping which take coarse to finer granularity features into consideration. Second, we introduce representative select module to get a more reliable matching result and matching score.

\subsection{Multi-order one-to-one mapping\label{section:mapping}}
Based on our multi-granularity feature, a group can be express by $G=(\mathcal{V}, \mathcal{E}, \mathcal{H}, \mathcal{A}, \mathcal{W}) $, which consists of nodes $v \in \mathcal{V}$, edges $e \in \mathcal{E}$ and hyper-edge $h \in \mathcal{H}$ with their attributes $a \in \mathcal{A}=(\mathcal{A}_{\mathcal{V}}, \mathcal{A}_{\mathcal{E}}, \mathcal{A}_{\mathcal{H}})$ and their weights $w \in \mathcal{W}=(\mathcal{W}_{\mathcal{V}}, \mathcal{W}_{\mathcal{E}}, \mathcal{W}_{\mathcal{H}})$. 
The goal of multi-order graph matching is to establish mapping between nodes of two multi-order graphs $G^{P}=(\mathcal{V}^{P}, \mathcal{E}^{P}, \mathcal{H}^{P}, \mathcal{A}^{P}, \mathcal{W}^{P})$ and $G^{Q}=(\mathcal{V}^{Q}, \mathcal{E}^{Q}, \mathcal{H}^{Q}, \mathcal{A}^{Q}, \mathcal{W}^{Q})$.

Suppose a set of all possible correspondence $\mathcal{C} = \mathcal{V}^{P} \times \mathcal{V}^{Q} $. The solution of multi-order matching is determined as a subset of candidate correspondences $\mathcal{C}$ and efficiently represented using a binary assignment matrix $\mathbf{X} \in \{0,1\}^{n^{P} \times n^{Q}} $, where $n^{P}$ and $n^{Q}$ denotes the number of nodes in $G^{P}$ and $G^{Q}$, respectively. If $v_{i}^{P} \in \mathcal{V}^{P} $ matched to $v_{a}^{Q} \in \mathcal{V}^{Q} $, $\mathbf{X}_{i,a}=1$, otherwise $\mathbf{X}_{i,a}=0$. In graph matching problem, it is necessary to impose two way constraint (i.e. one-to-one constraint) that make $\mathbf{X}$ a permutation matrix.

\begin{figure}[t!]
  \centering
  \includegraphics[width=0.45\textwidth]{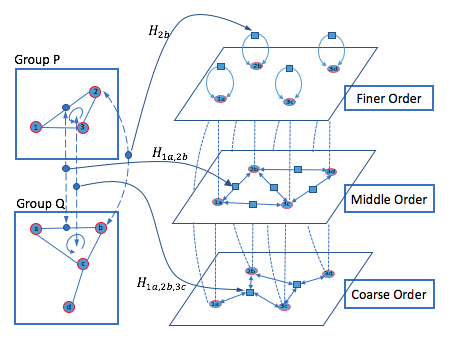}
  \caption{Multi-order association graph.}
    \label{fig:matching}
  \vspace{-5mm}
\end{figure}

To solve the multi-order matching problem in a random walk view, we propose a multi-order association graph structure $G^{M}=(\mathcal{V}^{M}, \mathcal{E}^{M}, \mathcal{H}^{M}, \mathcal{A}^{M}, \mathcal{W}^{M})$ to jointly describe multiple granularity while preserving characteristics of each granularity which is shown in Fig.~\ref{fig:matching}. The multi-order graph structure is inspired from the multi-layer association graph \cite{2016multi} that is used in multi-attribute graph matching. Our multi-order association graph structure consists of multiple orders to share the same indices set of matching candidates as described in Fig.~\ref{fig:matching}.  Each vertex $v_{ia}^{f} \in \mathcal{V}^{M}$ is different from the other vertices according to the order index (e.g. $v_{ia}^{f} \neq v_{ia}^{m}$) and can be connected to other vertices through two types of links: \emph{intra-order} and \emph{inter-order} links. 

{\bf The \emph{Intra-order} Transition Tensor Construct.} The \emph{intra-order} relation (i.e. relations between vertices which have the same order) are independently defined for each order. For Finer order, the relation from a vertex to itself $H_{ia}$ denotes the similarity of this matching candidates (i.e. $sim(v_{i}^{P},v_{a}^{Q})$). Similarly, for middle order, the relation between two matching candidates $H_{ia,jb}$ denotes the similarity of this matching edge candidate (i.e. $sim(e_{i_{1}i_{2}}^{P},e_{a_{1}a_{2}}^{Q})$). For coarse order, the relation among three matching candidate $H_{ia,jb,kc}$ denotes the similarity of this matching hyper-edge candidate (i.e. $sim(h_{i_{1}i_{2}i_{3}}^{P},h_{a_{1}a_{2}a_{3}}^{Q})$). Thus, we can construct a rank-3 affinity tensor $H$.

Intuitively, the same people/people-subgroup from different camera would be more likely to have similar importance value than those of different people/people-subgroups. Thus, the difference in importance score can be used as a penalty to the similarity score. Since large importance scores are used to enhance the similarity score of matched people/people-subgroups. We define rank-3 \emph{acceptance tensor} $\mathcal{W}$ by a bi-directional weighting mechanism and each element represent how contributive a matched objects/objects-subgroups candidate is.
\begin{align}
\mathcal{W}_{i_{1}a_{1},\cdots,i_{k}a_{k}}^{(k)}=\frac{w(x_{i_{1},\cdots,i_{k}}^{P})\cdot w(x_{a_{1},\cdots,a_{k}}^{Q})}{\alpha + |w(x_{i_{1},\cdots,i_{k}}^{P})-w(x_{a_{1},\cdots,a_{k}}^{Q})|}
\end{align}
where $\alpha$ is a parameter controlling the penalty of importance difference and prevent the denominator to be 0.

In order to separate the affinity of different granularity into different order. We define \emph{selection tensor} as follows:
\begin{align}
        \mathcal{I}^{k}(i_{1},i_{2},i_{3})
        =\left\{
           \begin{aligned}
           &0,~~~~~|unique(i_{1},i_{2},i_{3})|=k\\
           &1,~~~~~\text{otherwise}
           \end{aligned}
        \right.\,,
\end{align}
where $|unique(i_{1},i_{2},i_{3})|$ denotes how many different number is in set $\{i_{1},i_{2},i_{3}\}$.

Finally, the inter-order transition tensor can be expressed as:
\begin{align}
\label{equation:intra-order}
\mathbf{P}^{\alpha;\alpha;\alpha}=\frac{H \circ \mathcal{I}^{\alpha} \circ \mathcal{W}}{d_{max}^{\alpha}}
\end{align}
$$
s.t.~~ d_{max}^{\alpha}=\mathop{\max}_{i}  \mathbf{d}_{i}^{\alpha}=\sum_{i_{2},i_{3}}((H \circ \mathcal{I}^{\alpha} \circ \mathcal{W}) \otimes_{2} \mathbf{1} \otimes_{3} \mathbf{1})_{i}
$$
where the operator $\circ$ denotes the Hadamard product, operator $\otimes$ denotes the tensor product and $\mathbf{1}$ is all one vector. Note that the division in Eq.\ref{equation:intra-order} is element-wise and $\mathbf{d}_{i}^{\alpha}$ denotes corresponds to the sum of all similarity with weight for node $i$. Note that a tensor and a vector can be multiplied in different ways. In this paper, we use following notation:
\begin{align}
B=A \otimes_k V,~~~ B_{i_1,\cdots,i_{k-1},i_{k+1},\cdots,i_n}=\sum_{i_k}A_{i_1,\cdots,i_k,\cdots,i_n}V_{i_k}
\end{align}

{\bf The \emph{Inter-order} Transition Tensor Construct.} The \emph{inter-order} relation (i.e. relations between vertices which have different orders) indicates a random walker can jump between vertices from different orders.
The affinity values of inter-order connections are uniformly set to 1 because the relative importance of the edges is unknown. Therefore, the inter-order adjacency tensor $H^{\alpha_{1};\alpha_{2};\alpha_{3}}$ can be initially defined as an $N^P N^Q \times N^P N^Q \times N^P N^Q$ all-ones matrix. Then, the row-wise probabilistic normalization is applied to construct the inter-order transition matrix $P^{\alpha_{1};\alpha_{2};\alpha_{3}}$. However, since the proposed multi-order association graph structure does not consider the inter-order connections between vertices that have different indices, non-diagonal elements should be set to zero in the normalization process. Moreover, each order should have the same transition probability because the relative importance of the orders is also unknown without any prior knowledge. Finally, the inter-order transition matrix $P^{\alpha_{1};\alpha_{2};\alpha_{3}}$ can be defined as follows:
\begin{align}
\mathbf{P}^{\alpha_{1};\alpha_{2};\alpha_{3}}=\frac{1}{N^P N^Q L} H^{\alpha_{1};\alpha_{2};\alpha_{3}}=\frac{1}{N^P N^Q L}\mathcal{I}^1
\end{align}
$$
\text{s.t.}~~\alpha_{1} \neq \alpha_{2} ~\text{or}~ \alpha_{1} \neq \alpha_{3} ~\text{or}~ \alpha_{2} \neq \alpha_{3}
$$
where $\mathcal{I}^{1}$ is actually a $n^{P}n^{Q} \times n^{P}n^{Q} \times n^{P}n^{Q}$ identity tensor. Unfortunately, the above uniform transition probability distribution can decrease the distinctiveness among matching candidates in practical environments. To handle this problem, we employ an assumption that strongly connected vertices with others have more valuable and reliable information to propagate than other vertices. The assumption is reasonable because true correspondences usually organize a strongly connected cluster in practical graph matching applications. Moreover, this assumption is one of the theoretical bases of the spectral matching algorithm, which is frequently adopted in recent graph matching researches. Based on the assumption, we can design a weight vector using the degrees of intra-order connections for computing relative importance among vertices. Then, the reinforced inter-order transition tensor can be defined as follows:
\begin{align}
\label{equation:inter-order}
\mathbf{P}^{\alpha_{1};\alpha_{2};\alpha_{3}}=diag(\frac{\mathbf{d^{\alpha_{1}}}}{d_{max}^{\alpha_{1}}}) \circ \frac{\mathcal{I}^{1}}{n^{P}n^{Q}L}
\end{align}
$$
\text{s.t.}~~\alpha_{1} \neq \alpha_{2} ~\text{or}~ \alpha_{1} \neq \alpha_{3} ~\text{or}~ \alpha_{2} \neq \alpha_{3}
$$
where $diag(\mathbf{d}^{\alpha_{1}}/d_{max}^{\alpha_{1}})$ is a diagonal tensor that describes the normalized degrees of the corresponding vertices.

\begin{figure}[t!]
  \centering
  \includegraphics[width=0.5\textwidth]{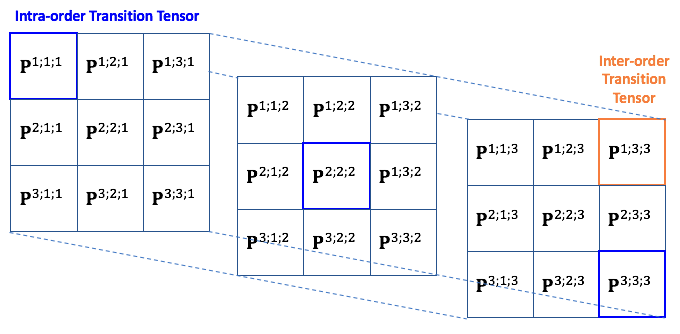}
  \caption{Illustration of supra-transition tensor.}
    \label{fig:tensorP}
  \vspace{-5mm}
\end{figure}

Naturally, a \emph{supra-transition tensor} $\mathbf{P}$ can be illustrate as Fig.~\ref{fig:tensorP}. Then, a multi-order graph matching problem can be formulated as:
\begin{align}
\hat{\mathbf{X}} &=\mathop{\arg \max}_{\mathbf{X}} F_{gm}(\mathbf{X}) =\mathop{\arg \max}_{\mathbf{X}}  \mathbf{P}\otimes_{1} \tilde{\mathbf{x}} \otimes_{2} \tilde{\mathbf{x}} \otimes_{3} \tilde{\mathbf{x}}
\end{align}
$$
\text{s.t.}~~\tilde{\mathbf{x}}=\mathbf{s} \otimes \mathbf{x},
\mathbf{X} \in \{0,1\} ^{n^{P} \times n^{Q}},
\mathbf{X}\mathbf{1}_{n^{Q}}<\mathbf{1}_{n^{P}},
\mathbf{X}^{T}\mathbf{1}_{n^{P}}<\mathbf{1}_{n^{Q}},
$$

where $\mathbf{s}$ is a 3-dimensional vector which describes relative confidence value among different order and $\mathbf{x}$ is the vectorized version of the matrix $\mathbf{X}$ for a convenient representation.

Then, we borrow the idea of hyper graph matching via reweighted random walk in \cite{RRWHM}, a random walker starts a traversal from any vertex and then randomly moves to a new vertex according to the supra-transition tensor. 
Suppose the current distribution of random walkers is given as $\tilde{\mathbf{x}}_{t}$, then the next distribution $\tilde{\mathbf{x}}^{t+1}$ can be described $P$ as follows:
\begin{align}
\tilde{\mathbf{x}}^{(t+1)}& = \mathbf{P} \otimes_{2} \tilde{\mathbf{x}}^{(t)} \otimes_{3} \tilde{\mathbf{x}}^{(t)} 
\end{align}

Random walkers traverse according to the transition tensor during the iteration steps, however, the traversal does not consider about the matching constraints. To impose the matching constraints on the random walks, we employ the personalized PageRand approach:
\begin{align}
\tilde{\mathbf{x}}^{(t+1)}& = \theta \mathbf{P} \otimes_{2} \tilde{\mathbf{x}}^{(t)} \otimes_{3} \tilde{\mathbf{x}}^{(t)} + (1-\theta)\mathbf{r}
\end{align}
where, r is the personalized vector and $\theta$ means a bias between random walking and personalized jumps. It means that the random walker travels along with its multi-order links with a probability $\theta$, or jumps according to the probability distribution $\mathbf{r}$ with a probability $1-\theta$. We impose the one-to-one constraints on the random walks using the reweighting jump $\mathbf{r}$. Adopting a reweighting function $f(\cdot)$, the reweighted random walks is formulated by:
\begin{align}
\label{equation:one-iter}
\tilde{\mathbf{x}}^{(t+1)}& = \theta \mathbf{P} \otimes_{2} \tilde{\mathbf{x}}^{(t)} \otimes_{3} \tilde{\mathbf{x}}^{(t)} + (1-\theta) f(\mathbf{P} \otimes_{2} \tilde{\mathbf{x}}^{(t)} \otimes_{3} \tilde{\mathbf{x}}^{(t)} )
\end{align}

Here, the first term denotes the affinity preserve random walk in the multi-order association network while the second term denotes the reweighting jump to meet the one-to-one constraint. The reweighted function $f(\cdot)$ is similarly to \cite{RRWHM} which consists of two steps: \emph{inflation} and \emph{bistochastic normalization}. 
The inflation step,  which is indicated in Line 9 of Algorithm~\ref{algorithm:matching}, attenuates small values and amplifies large values of distribution vector. In this way, unreliable correspondences contribute insignificantly through the individual exponentials over the components of distribution vector. Thus, unreliable matching candidates can be filtered out. Then, for the two-way constraint that a node in the graph $G^P$ must correspond to only one node in the graph $G^Q$ and vice versa, the bistochastic normalization scheme is applied in Line 10 of Algorithm~\ref{algorithm:matching}, which alternatively normalizes the rows and columns of $\mathbf{X}$ (matrix form of distribution vector). Note that any square matrix whose elements are all positive is proven to converge to a bistochastic matrix just by the iterative process. As a result, the contradictions among the matching candidates in the assignment distribution could be removed. These steps are separately applied for each order to preserve the distinctiveness of multiple orders.

After the reweighting process, the reweighted distribution vectors should be integrated to prevent a contradiction among the layers. One of the possible methods is gathering the distribution vectors according to any predefined priority of orders at once, and then propagating again to each order. However, it is difficult to decide the most important order without any prior knowledge, because different granularities have various characteristics which is hard to compare with. For that reason, we empirically define a order importance measure as the intersection of reweighted distribution and current assignment distribution. The proposed measure is based on the observation that the large difference between the current assignment distribution and the reweighted distribution means that the reweighting process causes large information loss. After calculating each order importance value, the importance vector s is normalized. At the end of iteration steps, the reweighted distribution is integrated with the current assignment distribution to generate the biased assignment distribution. Finally, the converged distribution vector is discretized to obtain a binary solution by using Greedy mapping or Hungarian method.

\subsection{Representative select\label{section:select}}
\begin{figure}[t!]
  \centering
  \includegraphics[width=0.5\textwidth]{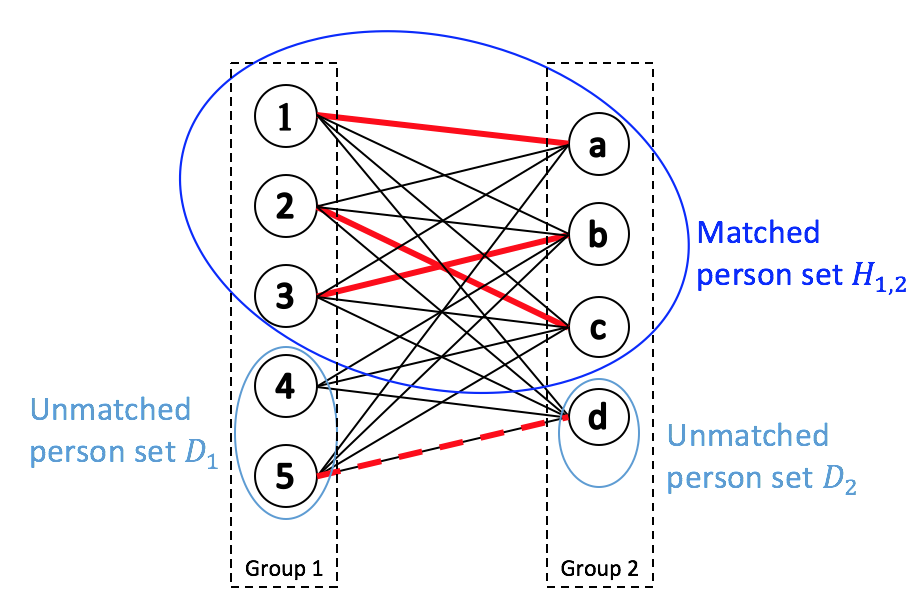}
  \caption{Example of matched and unmatched person sets (best viewed in color). The black and thin line indicates a possible match pair while the red and thick line indicates the best matches by our multi-order matching process. The red dashes indicates a matched pair excluded from matched set $H_{1,2}$.}
    \label{fig:unmatched}
  \vspace{-5mm}
\end{figure}

After we obtain one-to-one matching result, we are able to calculate the matching score. Furthermore, in order to avoid the interference from wrong person pair, person pair with small similarity value are also excluded from matched person set. Based on matched person set, we derive a new assignment vector $\hat{\mathbf{x}}$. Since the number of people may not the same in probe group and gallery group, we consider the person pair excluded from matched person set and person who do not have a match as the unmatched person set $D_{1}$ and $D_{2}$ (c.f. Fig.~\ref{fig:unmatched}). Thus, we define the similarity score function between two groups $P,Q$ from camera $A$ and $B$, respectively as follows:
\begin{align}
\label{equation:score}
score(P,Q)=&\sum_{\alpha} g(H\circ\mathcal{I}^{\alpha}\circ\mathcal{W}\circ\Pi)- \nonumber \\
&\sum_{v\in D_{1}\cup D_{2}}\frac{1}{1+exp(-Importance(v))}
\end{align}
$
\text{where}~~
        \Pi(i,j,k)
        =\left\{
           \begin{aligned}
           &0,~~~~~\hat{\mathbf{x}}_{i}=\hat{\mathbf{x}}_{j}=\hat{\mathbf{x}}_{k}=1\\
           &1,~~~~~\text{otherwise}
           \end{aligned}
        \right.\,,
$
and $g(\cdot)$ denotes the average of non-zeros value in the tensor. So the first term indicates the summation similarities in different granularities of matched person set, while the second term denote the penalty of unmatched person set. We will show that the penalty term of unmatched person set will improve the group re-identification significantly since a ``good'' group (i.e. a group with strong connection) is likely to have the same number of person in different camera view. Even if the number of person is not same, the unmatched person set is likely be a person with relatively low importance so that the penalty function will remove the interference of wrong group pair and guarantee the true group pair have a large matching score. The detail of our multi-order matching process is illustrated in Algorithm~\ref{algorithm:matching}.

Note that our matching process are different from many exist graph match method. In existing hyper-graph matching method \cite{RRWHM}, they embed finer granularity information (i.e. individual similarity) into coarse granularity (i.e. hyper-edge similarity). Therefore, the information from multiple granularity are not often fully exploit. Comparatively, our approach consider a multi-order structure which can make full use of multiple granularity information. We also different from multi-attributed graph matching method \cite{2016multi} in: (1) their approach  only separate the different attributes into different layers where granularity in each layer are the same, which can not be applied to match multi-granularity features. Comparatively, our multi-order matching process was designed to match multi granularity features. (2) Their approach assume every object has the same weight which have limitations in our group-wise matching case. We introduce importance to object/object-subgroups in different granularity based on the intuition that different people/people-subgroups carries different quantity of information. By allocating different importance and select most representative objects/object-subgroups in matching, it can handle interference of large appearance change and object variations, so as to obtain a more reliable matching result and matching score.

\begin{algorithm}[t]
   \caption{Multi-order Matching Process}
   \label{algorithm:matching}   
     {\bf Input}: Multi-order affinity tensor H, acceptance tensor $\mathcal{W}$, reweight factor $\theta$, inflation factor $\rho$\\
     {\bf Output}: Assignment vector $\hat{\mathbf{x}}$, matching score, F-score
     \begin{algorithmic}[1]
      \State \emph{(Initialization)}
      \State Generate an uniform assignment vector $\tilde{\mathbf{x}}$
      \State Generate an multi-order transition  tensor $P$ by Eq.~\ref{equation:intra-order} and Eq.~\ref{equation:inter-order}
      \Repeat
        \State \emph{(Calculate the next distribution)}
        \State $\tilde{\mathbf{x}} \leftarrow \mathbf{P} \otimes_{2} \tilde{\mathbf{x}} \otimes_{3} \tilde{\mathbf{x}}$ 
        \For {$\alpha=1 \to 3$}
          \State \emph{(Reweighting random walk for each order)}
          \State $\mathbf{u}^{\alpha} \leftarrow exp(\rho \cdot \tilde{\mathbf{x}}^{\alpha}/max(\tilde{\mathbf{x}}^{\alpha}))$ 
          \State Bistochasitic normalize $\mathbf{u}^{\alpha}$
          \State \emph{(Compute order importance)}
          \State $\mathbf{s}^{\alpha} \leftarrow sum(intersect(\mathbf{u}^{\alpha},\tilde{\mathbf{x}}^{\alpha}))$ 
        \EndFor
        \State Normalize the order importance vector $\mathbf{s}$
        \State \emph{(Gathering reweighted distribution with order confidence)}
        \State $\mathbf{u}^{temp} \leftarrow \sum_{\alpha} \mathbf{s}^{\alpha} \mathbf{u}^{\alpha}$
        \State \emph{(Diffusing reweighted distribution to whole order)}
        \State $\forall \alpha~\mathbf{u}^{\alpha} \leftarrow \mathbf{u}^{temp} $
        \State \emph{(Random walking with reweighted jumps)}
        \State $\tilde{\mathbf{x}} \leftarrow \theta \tilde{\mathbf{x}} + (1-\theta)\mathbf{u}$
      \Until {$\tilde{\mathbf{x}}$ converges.}
      \State \emph{(Integrate the assignment vector)}
      \State $\mathbf{x} \leftarrow \sum_{\alpha} \tilde{\mathbf{x}}^{\alpha}$
      \State Discretize $\mathbf{x}$ by the matching constraints
      \State Remove the matched pair with small similarity and get a new assignment vector $\hat{\mathbf{x}}$
      \State Calculate matching score and F-score
     \end{algorithmic}
\end{algorithm}

\section{Experimental Results\label{chapter:experimental_evaluation}}

\subsection{Datasets and Experimental Settings}
We perform experiments on the following three datasets:

{\bf MCTS.} The i-LID multiple-camera tracking scenario (MCTS) dataset \cite{zheng} was captured at an airport arrival hall in the busy times under a multi-camera CCTV network. Zheng et al.\cite{zheng} extracted image frames captured from two non-overlapping camera views and construct a group re-identification dataset which contains 64 groups and 274 group images. Most of groups have 4 images and every group contain 2 to 5 objects. The group layout of this dataset is rarely change, but it suffers from the low quality image and great illumination variation through different camera. In Zheng's paper, they randomly selected one image from each group to build the gallery set and the other group images formed the probe set. However, in our paper, we choose two images from different camera of each group and build the probe set and gallery set for unifying with other own constructed dataset.

{\bf DukeMTMC.} Recently, the DukeMTMC dataset \cite{2016MTMC} was released to quantify and evaluate the performance of multi-target, multi-camera tracking systems. It is high resolution 1080p, 60fps dataset and includes surveillance footage from 8 cameras with approximately 85 minutes of videos for each camera. There are cameras with both overlapping and non-overlapping fields-of-view. The dataset is of low density with 0 to 54 people per frame. We extract 177 group images pairs from different camera pair. It is challenging due to viewpoint changes since there are 8 cameras with significantly different view in the dataset. In the meantime, it is also challenging because of the large number of person (i.e. 3 to 10 people) and sheltering within a group.

{\bf Road.} The Road dataset is our own constructed dataset which includes 162 group pairs taken by two camera with camera $A$ monitoring an exit region and camera $B$ monitoring a road region.
Since images in this dataset are taken from a realistic crowd road scene at Tianmu Road, the interferences from severe occlusion and large layout variation significantly increase the difficulty of this dataset. The layout variation include shape variation, direction variation, compactness variation and amount variation. We include different situation in this dataset to demonstrate the effectiveness of our approach.

Note that we follow a classical sociological interpretation of groups: A group is defined as two or more people interacting to reach a common goal and perceiving a shared membership, based on both physical and social identity. To be specific, The physical identity can be regarded as a static relation connecting physical distance to group membership. Social identity reflects in the way people mutually influence each other and consequently move in groups, suggesting that social identity could be observed through trajectories shape similarity and paths temporal causality. Besides, people may merge in groups starting from different location (e.g. meeting action) or groups may split into subgroups and singletons (according to the hierarchical coherence property of group formation). So in our dataset construction, two group are defined as the same group when the number of intersection of two group are larger than the one forth of the total number of object in these two group.

For all of the above datasets, we randomly select $50\%$ of group pairs for testing and get the average of five tests since our method is in an unsupervised manner. Note that we use the extend version of \cite{2016groups} to extract groups in different cameras, and manually find group pair from different cameras. Then, we apply object detection method\cite{reinspect} to identify object in each group image and scaled them to $128\times 48$ for feature extraction. We perform evaluation in terms of standard Cumulative Matching Characteristic(CMC) curve~\cite{cmc} which  measures the correct match rates within different Re-ID rank ranges and F-Score\cite{group-info2} on 1-1 assignment which is defined as $2\times(\text{precision}\times \text{recall})/(\text{precision} + \text{recall})$.
 
{\bf Parameters:} In the group re-identification framework, we set $\alpha_{pur}=\alpha_{stb}=1$,$\sigma_r=\sigma_s=0.5$, maximum iteration number to be 5. In our matching process, we fix the reweight factor $\theta$ to 0.2, inflation factor $\rho$ to 30.

\subsection{Results for group re-identification framework}

We compare results of six methods on the Road dataset: (1) Simply utilize global features of whole group image for group Re-ID. In another word, we apply our finer granularity feature to the whole group image (\emph{Global}); (2) Only consider the finer granularity and apply our matching method for group Re-ID (\emph{Finer}); (3) Consider the finer and middle granularity, and then apply our matching method  (\emph{Finer+middle});  (4) Using proposed multi-granularity group Re-ID framework and multi-order matching process, but a poor detection when performing Re-ID (\emph{Poor-detection}); (5) Using proposed multi-granularity group Re-ID framework and multi-order matching process, but with groundtruth detection when performing Re-ID (\emph{Groundtruth}); (6) Multi-granularity Group Re-ID framework without assigning importance (i.e. every objects/object-subgroups are considered equally.) (\emph No-assign); (7) Our approach (\emph{Proposed}).



\begin{figure*}[t!]
  \centering
  \subfigure[]{\includegraphics[width=6cm]{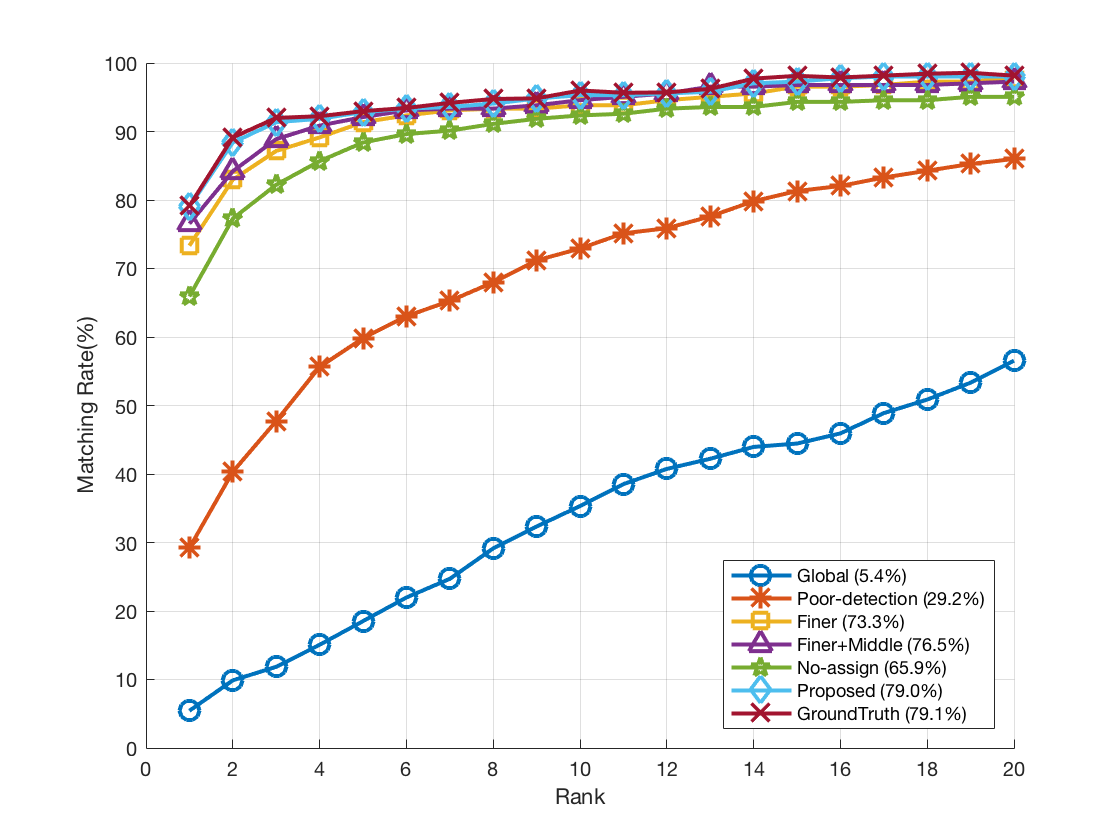}
    \label{fig:reid cmc curve a}}
    \hspace{-8mm}
  \subfigure[]{\includegraphics[width=6cm]{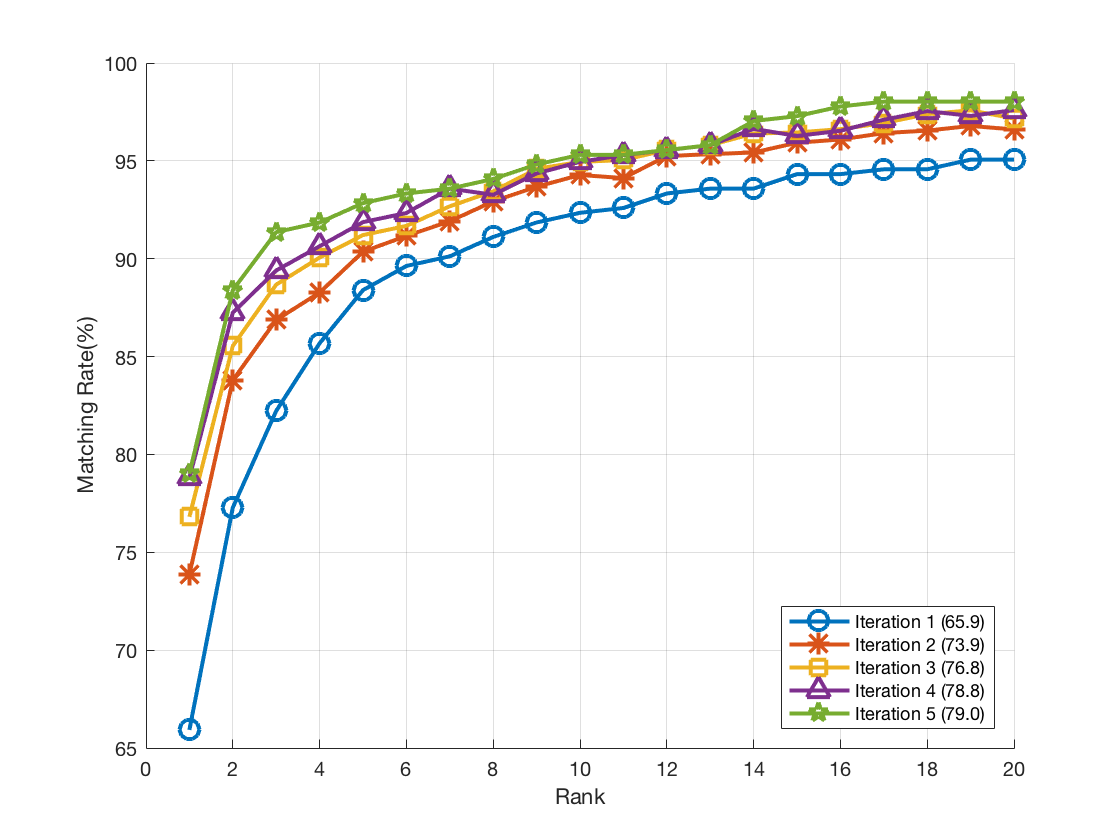}
    \label{fig:reid cmc curve b}}
    \hspace{-8mm}
  \subfigure[]{\includegraphics[width=6cm]{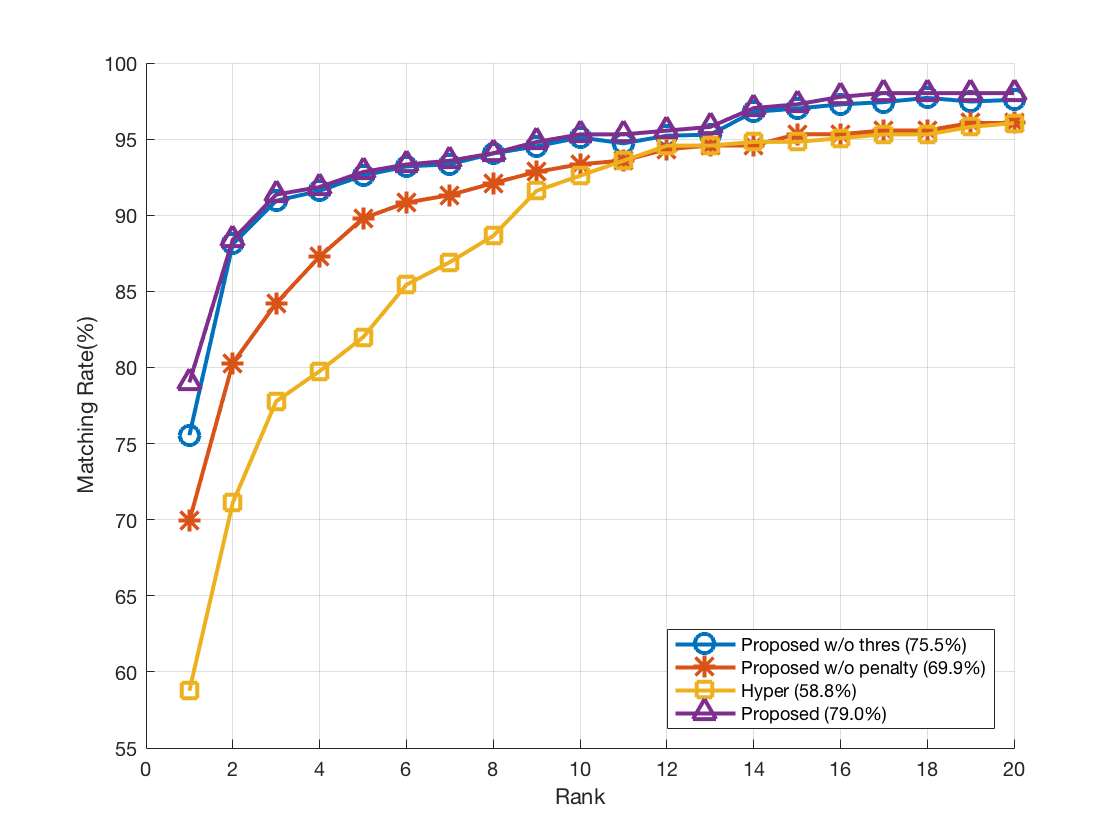}
    \label{fig:matching cmc curve}}
    \hspace{-8mm}
  \caption{CMC result on Road dataset. (a) CMC curves for different variant methods based on our group re-identification framework. This figure shows the effectiveness of our proposed group re-identification framework. (b) CMC curves of each iteration. This figure shows the improvement over iterations while learning the importance of different people/people-subgroups. (c) CMC curve curves for different variant methods based on our multi-order matching process.}
    \label{fig:reid cmc curve}
\end{figure*}

Fig.~\ref{fig:reid cmc curve a} shows the CMC results of different variant methods based on our group re-identification framework. From the CMC results, we can see that: 
\begin{enumerate}
\item[1)] Our approach has better Re-ID performances than other methods. This demonstrates the effectiveness of our approach. 
\item[2)] Our approach has obviously improved results than the \emph{Finer} method and \emph{Finer+middle} method. This indicates that multi-granularity features can effectively improve Re-ID performances by catching the most informative people/people-subgroups.
\item[3)] Our approach is better than the \emph{Poor-detection} method but has similar performance to the \emph{Groundtruth} method. This implies that a good detection result is necessary for group Re-ID. In contrast, a poor detection will decrease the group re-identification performance. 
\item[4)] Our approach also has superior performance than the \emph{No-assign} method. This demonstrates the usefulness of introducing importance of objects/object-subgroups in the matching process.
\end{enumerate}

Fig.~\ref{fig:reid cmc curve b} shows the CMC results of each iteration. We can see the CMC curve is improving by each iteration. This is because the importance of different people/people-subgroups become more and more accurate in the unsupervised learning process since the matching result and importance of each people/people-subgroups is interactive. After several iterations, the CMC curve become stalble which means the matching result become convergence.

\subsection{Results for multi-order matching process}

We compare the matching results of four methods: (1) Proposed method without the thresholding in Line 25 of Algorithm~\ref{algorithm:matching}, which means we do not exclude the matching pair with small affinity. (\emph{Proposed w/o threshold}). (2) Proposed method without the penalty term in score function Eq.\ref{equation:score} (\emph{Proposed w/o penalty}). (3) Using Hyper-graph matching method\cite{RRWHM} instead of our multi-order matching method (\emph{Hyper}); (4) Our approach which consists of a multi-order matching process and a representative select (\emph{Proposed}).

Fig.~\ref{fig:matching cmc curve} show the CMC results of different variant methods based on our multi-order matching process. From the CMC results, we can observe that: 
\begin{enumerate}
\item[1)] Without thresholding, the performance of group Re-ID is slightly decreased since it introduce the interference of wrong person pair. 
\item[2)] Our approach has obviously improved the performance than the \emph{Proposed w/o penalty} method. This indicates that people number variation is a important factor and the same group in more likely to have a small people number variation.  
\item[3)] Our approach also has superior performance than using hyper-graph matching method. This further demonstrates the effectiveness of the combination of our multi-granularity framework and multi-order matching process in group re-identification.
\end{enumerate}

\subsection{Compare with other methods}
\begin{figure*}[t!]
  \centering
  \subfigure[the MCTS dataset]{\includegraphics[width=6cm]{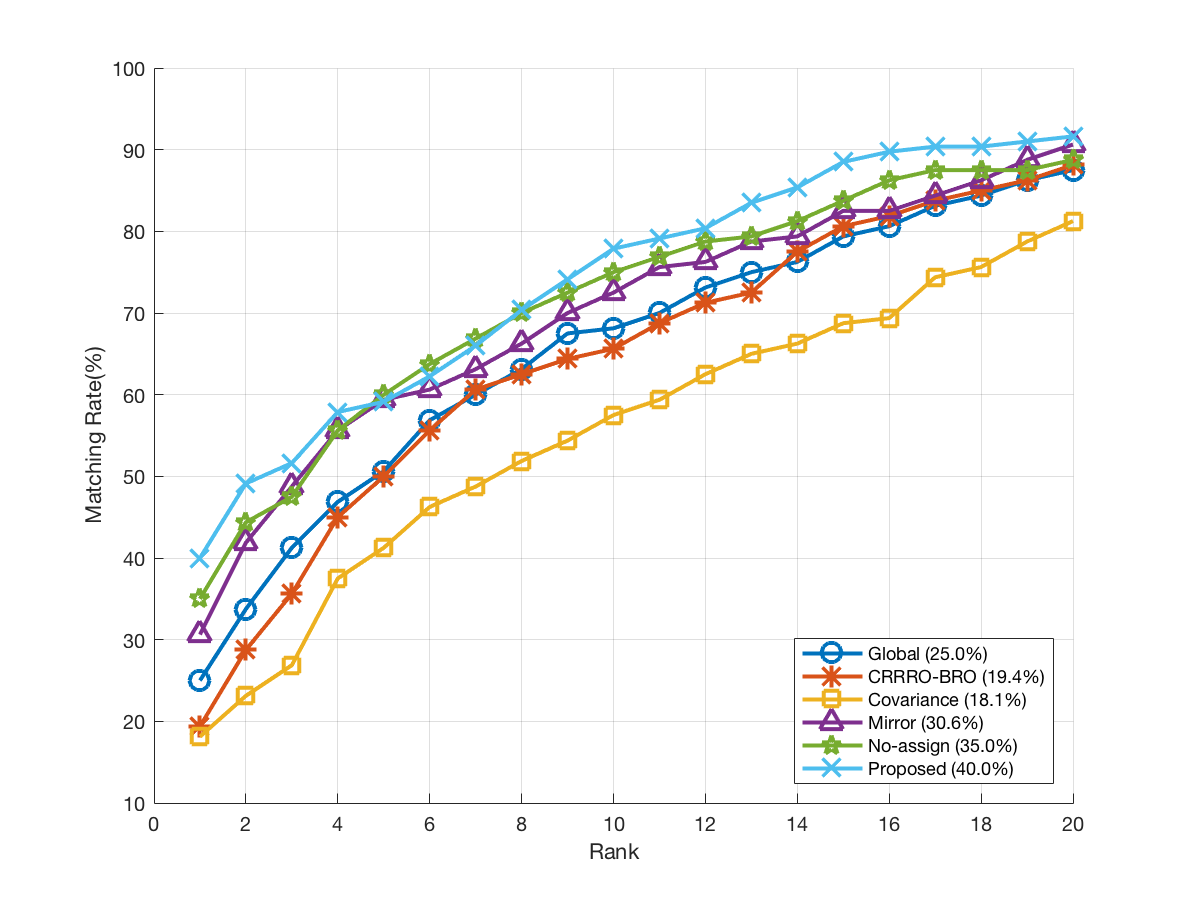}
    \label{fig:cmc curve a}}
    \hspace{-8mm}
  \subfigure[the DukeMTMC dataset]{\includegraphics[width=6cm]{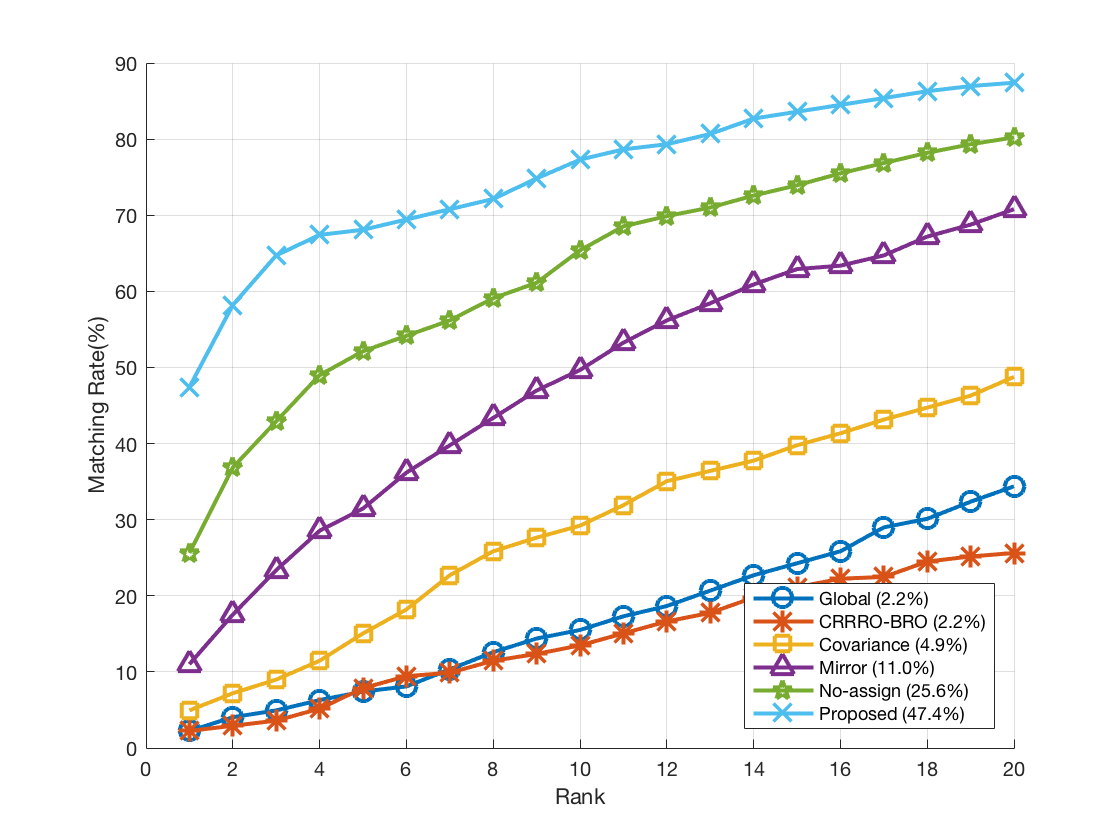}
    \label{fig:cmc curve b}}
    \hspace{-8mm}
  \subfigure[the Road dataset]{\includegraphics[width=6cm]{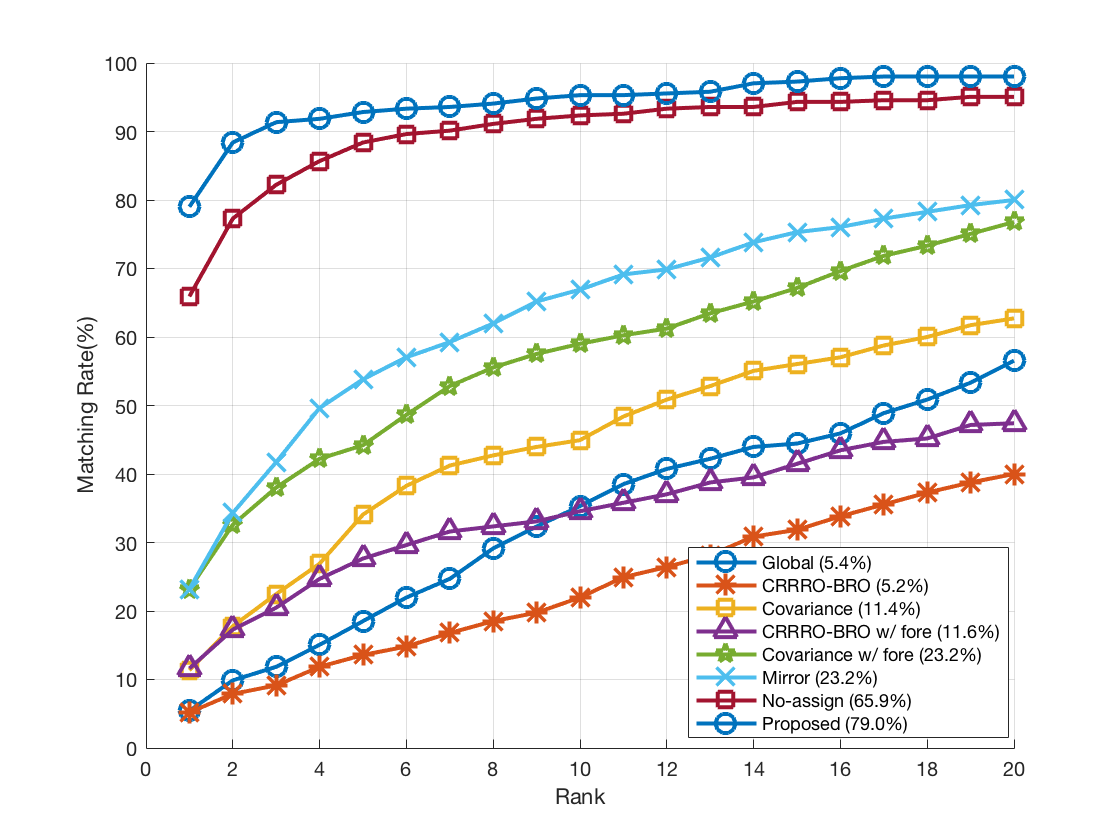}
    \label{fig:cmc curve c}}
  \caption{CMC curves for different methods on three dataset: MCTS, DukeMTMC, Road.}
    \label{fig:cmc curve compare}
\end{figure*}

Since group re-identification is seldom studied, only a few methods are proposed \cite{zheng,cai,zhu}. We compare our approach with global feature, CRRRO-BRO \cite{zheng} and Covariance descriptor \cite{cai}. CRRRO-BRO used a fixed template to segment group image into multiple ring regions and blocks, then proposed the Center Rectangular Ring Ratio-Occurrence Descriptor (CRRRO) and Block based Ratio-Occurrence Descriptor (BRO) to capture the the ratio information within and between regions. Cai \cite{cai} proposed a discriminative covariance descriptor to capture both appearance and statistic properties of image region to maintain illumination invariant. Both of these two method manually designed distance function. Since these two method both calculate features based on pixels, the performance should be improved if we only consider the foreground region. Thus, in our Road dataset, background subtraction is performed. We further design metric learning method based on our global features. To be specific, Mirror representation \cite{mirror} of Marginal Fisher Analysis (MFA) to evaluate global features with learned distance metric. The result is shown in Fig.~\ref{fig:cmc curve compare} and Table~\ref{tab:cmcTable1} to \ref{tab:cmcTable3}. 

\begin{table*}
\centering
\caption{CMC results on the MCTS dataset}\label{tab:cmcTable1}
\footnotesize{
\begin{tabular}{|p{3cm}|c|*{5}{c|}c|}
\hline
\textbf{Method}& Rank-1& Rank-5& Rank-10& Rank-15& Rank-20& Rank-30& F-score\\
\hline
Global& 25\% & 50.6\% & 68.1\% & 79.4\% & 87.5\%  & 98.7\% & - \\
CRRRO-BRO\cite{zheng}& 14.3\% & 43.1\% & 68.7\% & 75.0\% & 80.6\% & 98.7\% & - \\
Covariance\cite{cai}& 18.1\% & 41.2\% & 57.5\% & 68.8\% & 81.3\% & 99.4\% & - \\
Mirror\cite{mirror}& 30.6\% & 59.4\% & 72.5\% & 82.5\% & 90.6\% & 98.8\% & - \\
No-assign& 35.0\%  & 60.0\% & 75.0\% & 83.8\% & 88.8\% & 96.9\% & 81.2\% \\
{\bf Proposed}& {\bf 39.0\% }& {\bf 58.1\% }& {\bf 76.9\% }& {\bf 87.5\% }& {\bf 90.6\% }& {\bf 100\% }& {\bf 83.2\% }\\
\hline
\end{tabular}}
\end{table*}

\begin{table*}
\centering
\caption{CMC results on the DukeMTMC dataset}\label{tab:cmcTable2}
\footnotesize{
\begin{tabular}{|p{3cm}|c|*{5}{c|}c|}
\hline
\textbf{Method}& Rank-1& Rank-5& Rank-10& Rank-20& Rank-30& Rank-50& F-score\\
\hline
Global& 2.2\% & 7.4\% & 15.5\% & 34.4\% & 48.1\% & 77.1\% & - \\
CRRRO-BRO\cite{zheng}& 2.2\% & 7.8\% & 13.5\% & 25.6\% & 37.8\% & 61.3\% & - \\
Covariance\cite{cai}& 4.9\% & 15.1\% & 29.2\% & 48.8\% & 59.3\% & 73.9\% & - \\
Mirror\cite{mirror}& 11.0\% & 31.4\% & 49.6\% & 70.8\% & 81.1\% & 92.8\% & - \\
No-assign& 25.6\% & 52.1\% & 65.4\% & 80.2\% & 86.5\% & 94.8\% & 76.6\% \\
{\bf Proposed}& {\bf 47.4\% }& {\bf 68.1\% }& {\bf 77.3\% }& {\bf 87.4\% }& {\bf 91.9\% }& {\bf 98.6\% }& {\bf 79.0\% }\\
\hline
\end{tabular}}
\end{table*}

\begin{table*}
\centering
\caption{CMC results on the Road dataset}\label{tab:cmcTable3}
\footnotesize{
\begin{tabular}{|p{3cm}|c|*{5}{c|}c|}
\hline
\textbf{Method}& Rank-1& Rank-5& Rank-10& Rank-20& Rank-30& Rank-50& F-score\\
\hline
Global& 5.4\% & 18.5\% & 35.3\% & 56.5\% & 72.6\% & 86.4\% & - \\
CRRRO-BRO\cite{zheng}& 5.2\% & 13.6\% & 22.0\% & 40.0\% & 53.6\% & 78.0\% & - \\
Covariance\cite{cai}& 11.35\% & 34.1\% & 44.9\% & 62.7\% & 74.6\% & 84.2\% & - \\
CRRRO-BRO w/ fore& 11.6\% & 27.6\% & 34.6\% & 47.4\% & 59.2\% & 77.3\% & - \\
Covariance w/ fore& 23.2\% & 44.2\% & 59.0\% & 76.8\% & 83.5\% & 91.6\% & - \\
Mirror\cite{mirror}& 23.2\% & 53.8\% & 66.9\% & 80.0\% & 85.7\% & 93.3\% & - \\
No-assign& 65.9\% & 88.4\% & 92.3\% & 95.1\% & 97.7\% & 99.7\% & 94.0\% \\
{\bf Proposed}& {\bf 79.0\% }& {\bf 92.8\% }& {\bf 95.3\% }& {\bf 98.0\% }& {\bf 98.0\% }& {\bf 100\% } & {\bf 94.2\% }\\
\hline
\end{tabular}}
\end{table*}

From Fig.~\ref{fig:cmc curve compare} and Table~\ref{tab:cmcTable1} to \ref{tab:cmcTable3} above, we can conclude that: 
\begin{enumerate}
\item[1)] Our approach has the best re-identification performance on different dataset. Notice that \emph{Mirror} method improve the performance than \emph{Global} method because the \emph{Mirror} method evaluate global features with learned distance metric. This improvement is obvious in our Road dataset, but the result is still has a large gap with our proposed method. Besides, the performance \emph{CRRRO-BRO} and \emph{Covariance} method can also be improved if we only consider the foreground pixel but the result is still far away from our proposed method. 
\item[2)] Our approach has obviously improved results than the no-assign method. This demonstrates the effectiveness of our multi-granularity group re-identification framework. Meanwhile, the F-score which evaluates the 1-1 assignments performance has also be improved. This indicate that we can find a more accurate mapping across different camera by introducing introducing importance of objects/object-subgroups in the matching process. 
\item[3)] Our approach has the most significant improvement on DukeMTMC dataset and Road dataset, but a small improvement on MCTS dataset. This is because the rich multi-granularity information on DukeMTMC dataset and Road dataset since the average people number on these two dataset is 3.4, 3.8 respectively, while 2.3 on the MCTS dataset.
\end{enumerate}

\section{Conclusion}

In this paper, we propose a novel framework for addressing a rarely-studied problem of group re-identification which aims to re-identify group of people from different cameras. Our framework consists of two key ingredients: 1) introducing a multi-granularity group re-identification process, which derives features for multi-granularity objects (people/people-subgroups) within a group and iteratively evaluates their importances during group Re-ID, so as to handle group-wise misalignments due to viewpoint change and group dynamics; 2) a multi-order matching process to select representative people/people-subgroups adaptively and integrating multi-granularity information from these objects/object-subgroups to obtain group-wise matching.

Under this framework, our future work is devoted to explore new variants of the two components, such as: 1) Since the performance of our framework can be greatly affected by the accuracy of pedestrian detection. If the detection model is not fine-tuned on a specific scene, the information of multi-granularity will decrease while the global feature will play a significant role in group re-identification. So we can fuse global information into our framework and get a more reliable result on different scenes. 2) In the iteration process, the exact convergence of our learning process is difficult to analyze since the matching result and importance of each people/people-subgroups is interactive. We find out the one-to-one matching result become stable within 5 iterations in our experiments, while the analysis of convergence is not proven. Thus, we need to give a scientific explanation of the convergence. (3) The complexity of our algorithm is $O(n^2)$, which is far from practice. We need to further accelerate our algorithm by reducing the complexity.



\bibliographystyle{IEEEtran}
\bibliography{reference}

\end{document}